\documentclass[11pt]{article}
\usepackage{amsmath, amssymb}
\usepackage{graphicx}
\usepackage{hyperref}
\usepackage{caption, subcaption}
\usepackage{booktabs, multirow, makecell, array}
\usepackage{tikz}
\usetikzlibrary{shapes.geometric, arrows.meta, positioning}
\usepackage{color,xcolor,colortbl}
\usepackage{listings}

\usepackage{cuted}     

\usepackage{soul}
\usepackage{pgfplots}
\pgfplotsset{compat=1.18}
\usepackage[ruled,vlined,linesnumbered]{algorithm2e}
\SetKwComment{Comment}{$\triangleright$\ }{}
\SetNlSty{textbf}{\{}{\}}
\usepackage[most]{tcolorbox}
\usepackage{pifont}
\usepackage{tcolorbox}
\tcbuselibrary{listings, breakable}
\usepackage{listings}
\usepackage{xcolor}
\definecolor{darkred}{HTML}{D1191F} 
\definecolor{darkgreen}{HTML}{04bf29}
\lstdefinelanguage{json}{
    basicstyle=\ttfamily\small,
    showstringspaces=false,
    breaklines=true,
    frame=none,
    literate=
     *{0}{{{\color{black}0}}}{1}
      {1}{{{\color{black}1}}}{1}
      {2}{{{\color{black}2}}}{1}
      {3}{{{\color{black}3}}}{1}
      {4}{{{\color{black}4}}}{1}
      {5}{{{\color{black}5}}}{1}
      {6}{{{\color{black}6}}}{1}
      {7}{{{\color{black}7}}}{1}
      {8}{{{\color{black}8}}}{1}
      {9}{{{\color{black}9}}}{1}
      {:}{{{\color{black}{:}}}}{1}
      {,}{{{\color{black}{,}}}}{1}
      {\{}{{{\color{black}{\{}}}}{1}
      {\}}{{{\color{black}{\}}}}}{1}
      {[}{{{\color{black}{[}}}}{1}
      {]}{{{\color{black}{]}}}}{1},
    string=[s]{"}{"},
}
\usepackage{enumitem}
\setlist[itemize]{leftmargin=*, topsep=0pt, itemsep=2pt, parsep=0pt}

\renewcommand{\cite}{\citet}

\tikzstyle{process} = [rectangle, rounded corners, minimum width=3cm, minimum height=1.2cm, text centered, draw=black, fill=blue!10]
\tikzstyle{decision} = [diamond, draw=black, fill=gray!20, text centered, inner sep=1pt, aspect=2]
\tikzstyle{startstop} = [ellipse, draw=black, fill=gray!30, minimum width=3.2cm, minimum height=1.2cm, text centered]
\tikzstyle{arrow} = [thick,->,>=Stealth]

\usepackage[final]{acl}

\usepackage{times}
\usepackage{latexsym}

\usepackage[T1]{fontenc}

\usepackage[utf8]{inputenc}

\usepackage{microtype}
\usepackage{float}
\usepackage{inconsolata}

\usepackage{graphicx}

\usepackage{etoolbox}
\usepackage{risys-title}

\RisysTitle{SAFIRE \includegraphics[height=1.6ex]{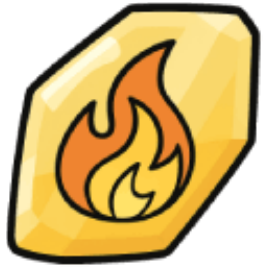}: Safety-Critical Benchmark for Fine-grained Fire and Smoke Understanding in Multimodal LLMs}
\RisysAuthor[]{Pengfei Li$^{1}$}
\RisysAuthor[]{Naufal Suryanto$^{1}$}
\RisysAuthor[]{Sicheng Zhang$^{1}$}
\RisysAuthor[]{Mohammad Alsharid$^{1}$}
\RisysAuthor[]{\\Muzammal Naseer$^{1,2}$}

\RisysAffil[]{Khalifa University$^{1}$}
\RisysAffil[]{\\University of Western Australia$^{2}$}
\RisysProjectpage{https://risys-lab.github.io/SAFIRE/}

\hypersetup{
    pdftitle={SAFIRE: Safety-Critical Benchmark for Fine-grained Fire and Smoke Understanding in Multimodal LLMs},
    pdfauthor={Pengfei Li, Naufal Suryanto, Sicheng Zhang, Mohammad Alsharid, Muzammal Naseer},
    pdfsubject={Safety-critical fire and smoke understanding benchmark for multimodal large language models},
    pdfkeywords={fire and smoke understanding, multimodality, benchmarking, cross-modal application, safety and alignment}
  }

\begin{document}

\RisysAbstract{Multimodal Large Language Models (MLLMs) show strong progress on vision--language tasks, yet their reliability in safety-critical settings remains underexplored. Fire-smoke understanding is central to public safety and disaster response, but most existing benchmarks lack diverse real-world scenarios and context-aware evaluation. We introduce SAFIRE, a large-scale benchmark for fire-smoke understanding in MLLMs, comprising 83K captioned images from 20 scenarios and 193K multiple-choice VQA (MCVQA) generated from a 9.7K-image subset, spanning 10 evaluation dimensions from basic perception to higher-order reasoning. A GPT-5.4-assisted multi-stage verification pipeline with MLLM majority voting ensures annotation quality. Evaluating ten open-source MLLMs (8B--38B) yields an average accuracy of 61.9\%, exposing major gaps in safety-critical reasoning. We further show that adapting vision encoders with only 7\% of our domain-specific data boosts fire-scene classification accuracy from 20.1\% to 64.5\%, indicating that carefully curated data can yield substantial gains even when data volume is limited.}

\RisysMakeTitle


\section{Introduction}


Fire and smoke occur across diverse human and natural activities such as cooking, industry, celebrations, accidents, and wildfires. While their visual patterns often appear similar across settings \cite{smoke2022survey}, the underlying causes, risks, and appropriate responses vary widely. Understanding these phenomena requires not only visual recognition but also reasoning about context, intent, and consequence, which goes beyond conventional detection or segmentation. Nonetheless, most existing studies remain limited to narrow recognition tasks based on small, privately curated datasets with minimal contextual annotation \cite{fsdtechnical, zhu2025multiscale_det,smokeseg,du2025firemultiformer,boumaraf2025vision}.

\begin{figure}[t!]
    \centering
    \includegraphics[width=0.95\columnwidth]{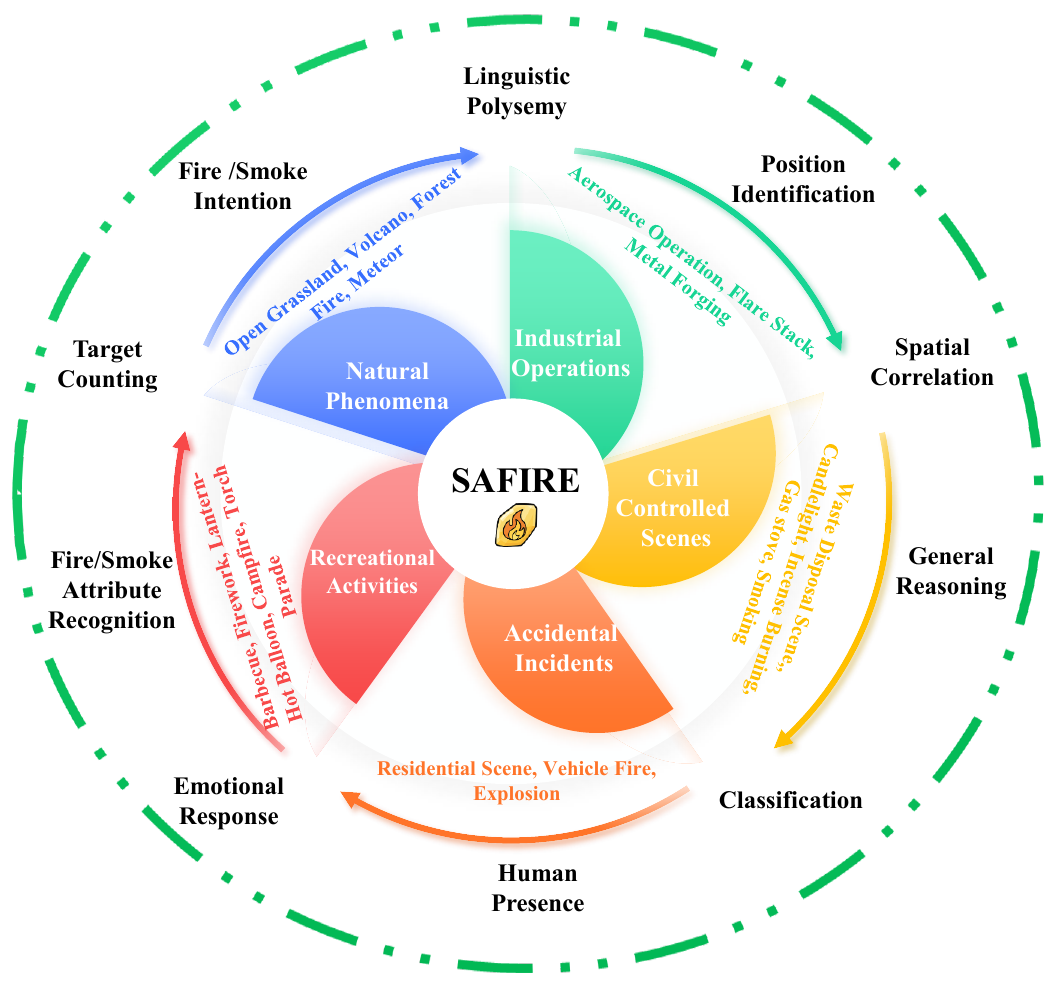}
    \caption{\textbf{From windmill center to the periphery:} SAFIRE consists of 5 processes spanning 20 real-world scenarios, with 10 evaluation dimensions and over 193K curated question-answer pairs derived from our subsets.
    }
    \label{fig:both}
    \vspace{-3.75mm}
\end{figure}

\begin{figure*}[!t]
    \centering
    \begin{minipage}[t]{0.49\textwidth}
        \centering
        \begin{subfigure}[t]{0.49\linewidth}
            \centering
            \includegraphics[width=\linewidth]{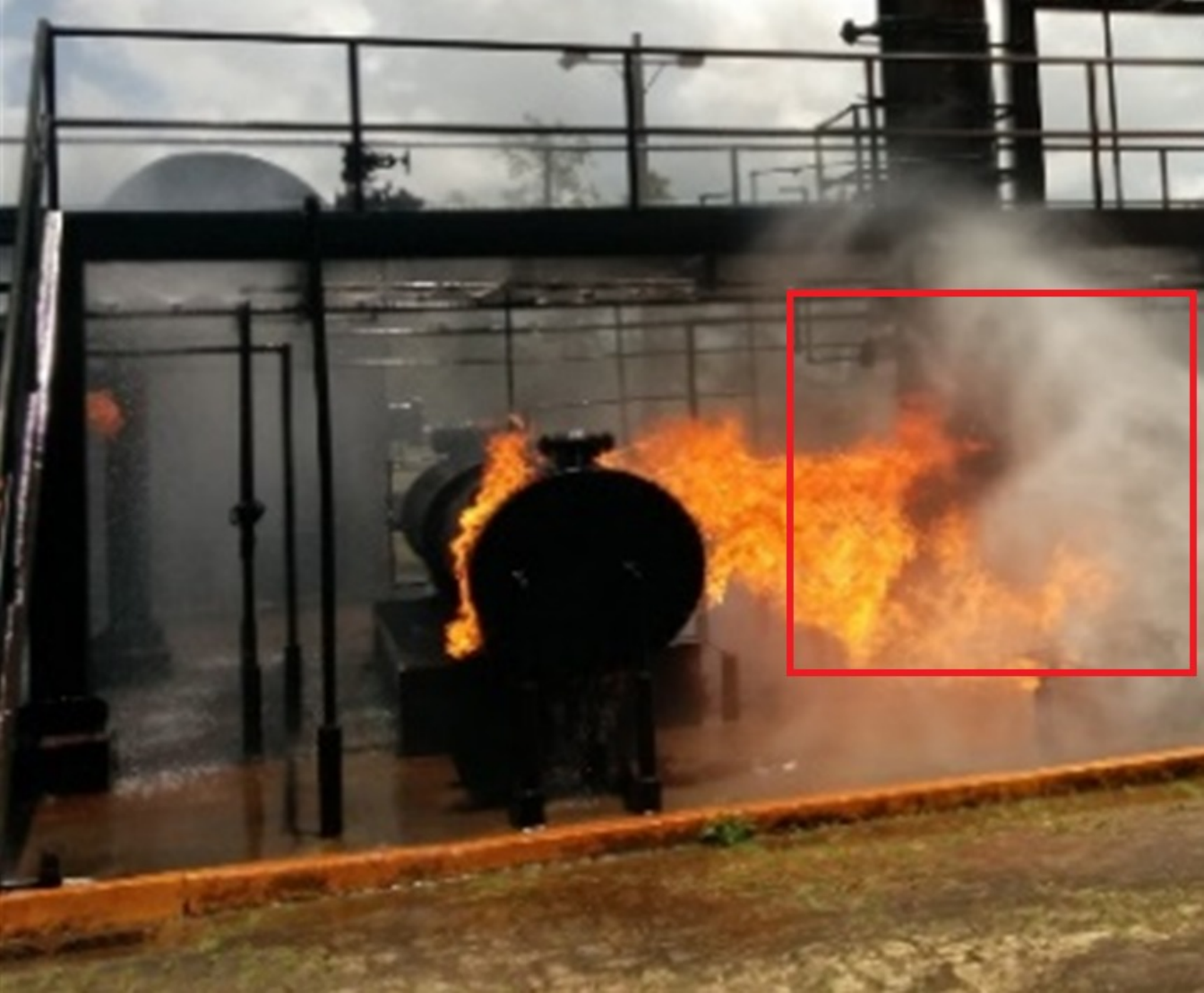}
            \caption{Misleading ``White Smoke"}
            \label{ff1}
        \end{subfigure}
        \hfill
        \begin{subfigure}[t]{0.49\linewidth}
            \centering
            \includegraphics[width=\linewidth]{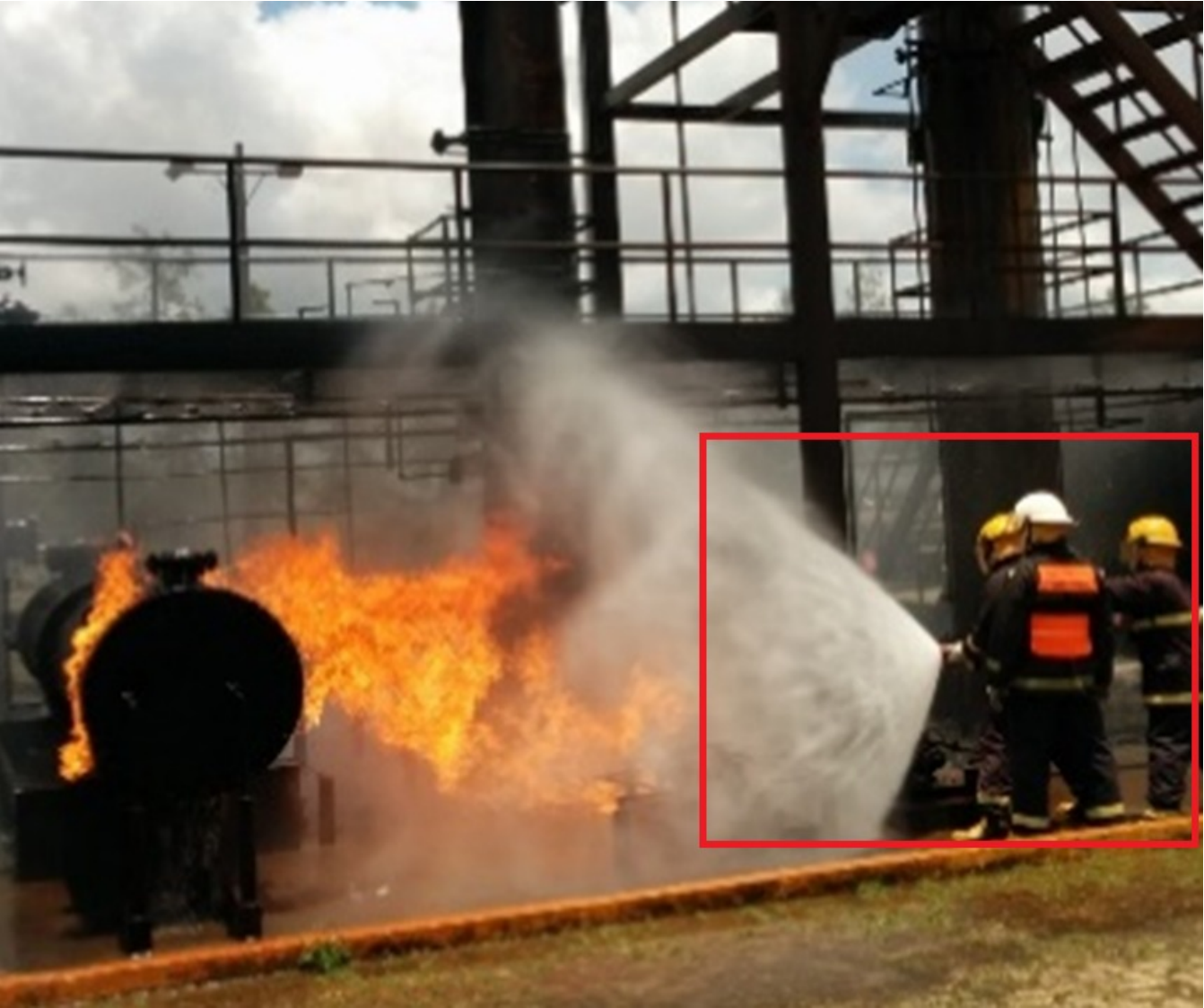}
            \caption{Recognizing “No Smoke"}
            \label{ff2}
        \end{subfigure}
        \caption{
        \textbf{Comparison of traditional and our dataset:}
        (a) False-positive “White Smoke” class image example from a local-perceptron dataset;
        (b) An image labeled ``No Smoke" in our dataset, with a caption recognizing firefighters and a fire hose reel.}
        \label{firefighting}
    \end{minipage}
    \hfill
    \begin{minipage}[t]{0.49\textwidth}
        \centering
        \begin{subfigure}[t]{0.49\linewidth}
            \centering
            \includegraphics[width=\linewidth]{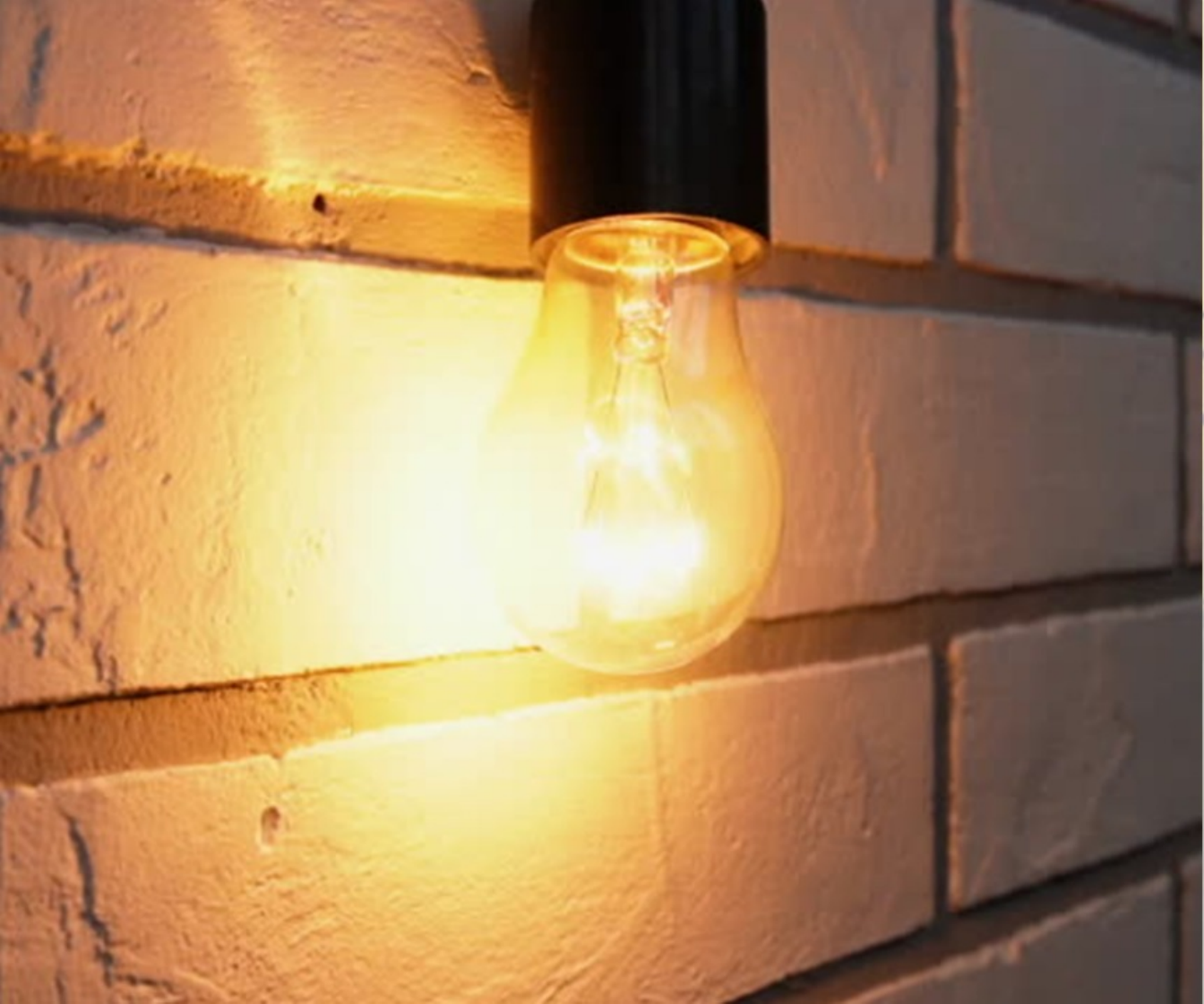}
            \caption{Light Bulb on Wall}
            \label{ff3}
        \end{subfigure}
        \hfill
        \begin{subfigure}[t]{0.49\linewidth}
            \centering
            \includegraphics[width=\linewidth]{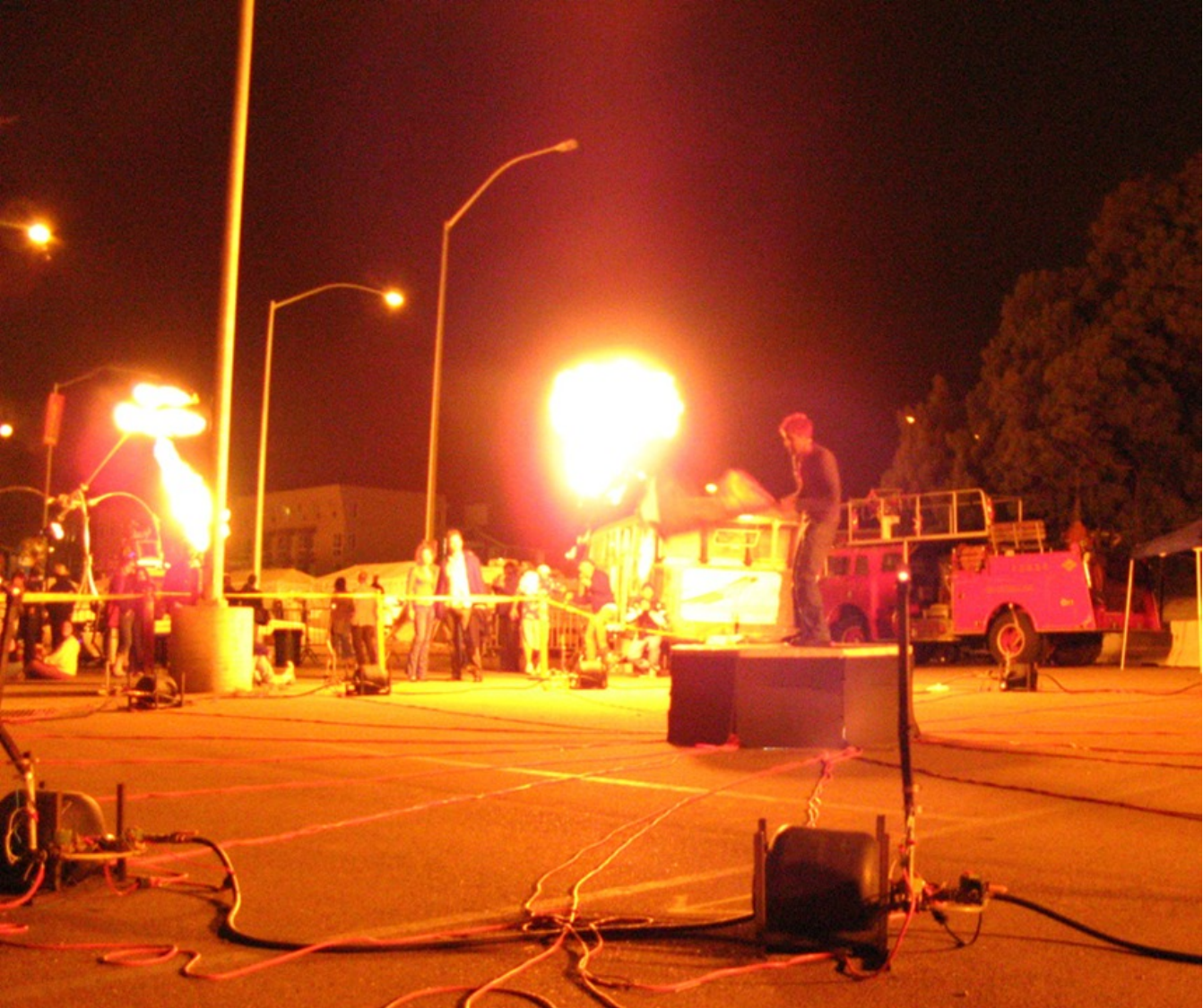}
            \caption{Stage Fire Performance}
            \label{ff6}
        \end{subfigure}
        \caption{
        \textbf{Misclassifications without contextual reasoning:}
        (a) A warm-toned bulb triggers a false-positive “fire”;
        (b) A live fire show yields a false-negative “no fire.”
        Incorporating global context, like identifying stage or fire trucks, corrects these mistakes.
        }
        \label{false_pos_neg}
    \end{minipage}
\end{figure*}

Recent MLLMs demonstrate strong zero-shot performance and richer contextual reasoning than traditional models \cite{liu2023visual,bai2025qwen25vltechnicalreport}, yet existing fire-smoke benchmarks fall short in evaluating the situational reasoning required in safety-critical contexts (Table~\ref{tab:compare_dataset}). They lack complex, context-aware queries such as determining whether visible smoke signals an emergency, leaving the reasoning limitations of current MLLMs insufficiently explored.

Figures~\ref{firefighting} and \ref{false_pos_neg} illustrate typical failure cases when contextual cues are ignored. In Fig.~\ref{firefighting}, water vapor is often misidentified as smoke during firefighting; in Fig.~\ref{false_pos_neg}(a), warm lighting from bulbs or streetlights may trigger false fire alarms; and in Fig.~\ref{false_pos_neg}(b), controlled pyrotechnic displays may appear as non-critical scenes. Broader scene cues such as firefighters, hoses, stages, or fire trucks help models resolve these ambiguities.

These challenges underscore the need for context-aware, multimodal reasoning in fire and smoke understanding, yet progress remains limited by the absence of large-scale, high-quality datasets aligned with modern MLLM capabilities. To bridge this gap, we introduce \textbf{SAFIRE} (\textbf{S}afety-\textbf{A}ware \textbf{F}ire-smoke \textbf{I}nspection and \textbf{R}easoning \textbf{E}valuation), a large-scale benchmark for comprehensive, context-aware evaluation of MLLMs in fire-smoke reasoning. As shown in Fig.~\ref{fig:both}, SAFIRE comprises 83K captioned images from 20 real-world scenarios and 193K multiple-choice VQA pairs (from a 9.7K subset) spanning 10 evaluation dimensions, supporting both perceptual and higher-order reasoning at the scene level. Our main contributions are summarized as follows:
\begin{itemize}[leftmargin=*]
\item We constructed \textit{SAFIRE}, a multi-scenario benchmark explicitly designed to evaluate MLLMs’ reasoning in fire-smoke contexts. Annotation quality is maintained through a GPT-5.4-assisted semantic verification pipeline combined with model-guided majority voting. 

\item \textit{SAFIRE} enables systematic assessment of scene-level, contextual, causal, and behavioral reasoning, providing a testbed that extends beyond conventional classification or detection tasks. Evaluations on ten open-source MLLMs yield an average MCVQA accuracy of 61.9\%, highlight remaining gaps in safety-critical reasoning.




\end{itemize}

\section{Related work}

\textbf{Classical fire/smoke datasets.}\quad
Early studies relied on handcrafted features and small binary datasets such as \textit{BowFire}~\cite{chino2015bowfire} and \textit{F.~Yuan}~\cite{yuan2015real}, which focused on color and texture cues to distinguish fire or smoke from background scenes. With the advent of deep learning, larger labeled datasets (e.g., \textit{Khan}~\cite{khandataset}, \textit{L.~He}~\cite{LHe}, \textit{EdgeFireSmoke}~\cite{almeida2022edgefiresmoke}) enabled convolutional and attention-based models. Early datasets such as \textit{FD}~\cite{FD} and \textit{MAFire-Net}~\cite{MAFire} adopted binary fire/no-fire setups, later evolving toward finer-grained classification. Notably, \textit{DFAN}~\cite{DFAN} introduced scene-specific categories (e.g., vehicle- or building-related fires), though with limited per-class samples and semantic overlap, while \textit{UFS-data}~\cite{hosseini2022ufs} proposed multi-label annotations for flame, white smoke, and black smoke. Despite these advances, video-derived frames and narrow scenario coverage restrict contextual diversity.


\noindent\textbf{Detection, segmentation, and richer labels.}\quad 
Recent benchmarks have increasingly emphasized fire--smoke localization and multi-scenario annotations to enhance model robustness. Datasets such as \textit{FiSmo}~\cite{fismo}, \textit{Nemo}~\cite{yazdi2022nemo}, \textit{DFS}~\cite{DFS}, and \textit{HQFSD}~\cite{hqfsd} establish foundational localization pipelines, while newer benchmarks like \textit{SmokeBench}~\cite{qi2026smokebench_wacv} and \textit{DetectiumFire}~\cite{liu2026detectiumfire} enrich these with bounding-box annotations, fine-grained severity taxonomies, and curated ``fire-like'' negative samples to rigorously evaluate detection accuracy and visual reasoning. To mitigate the visibility degradation that often compromises fire smoke perception tasks, \textit{SmokeBench$_{\mathrm{(MM)}}$}~\cite{jin2025smokebench_mm} contributes nearly 10K precisely aligned paired images specifically designed for early-stage fire desmoking, serving as a preprocessing foundation for downstream analysis. Furthermore, segmentation and sensor-augmented datasets have expanded the research landscape: \textit{FireSentry}~\cite{zhou2026firesentry} delivers synchronized infrared--visible video streams and pixel-level fire masks for fine-grained wildfire spread forecasting, while resources like \textit{WildfireSpread}~\cite{gerard2023wildfirespreadts} and \textit{UniInd-FireSmoke}~\cite{UniInd} incorporate satellite or industrial sensor modalities. 


\begin{table}[!t]
\centering
\caption{\textbf{Comparison of SAFIRE with Previous Influential Works.} ``Influential works'' here refers to widely cited historical benchmarks (2015-2020) or foundational recent benchmarks (2021-2026).
\includegraphics[height=0.9em]{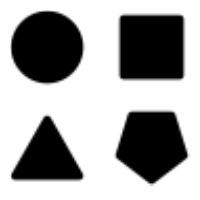} denotes Classification, \includegraphics[height=0.9em]{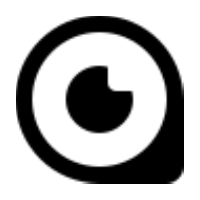} denotes Detection,
\includegraphics[height=0.9em]{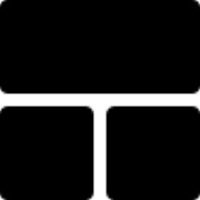} denotes Segmentation, \includegraphics[height=0.9em]{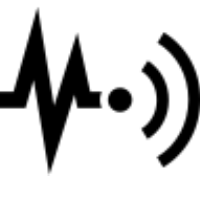} denotes Sensor Data,
\includegraphics[height=0.9em]{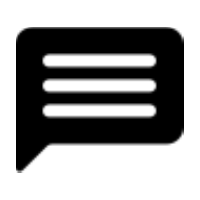} denotes MCVQA and/or Captions. The proposed SAFIRE benchmark stands out for its overall scale, scenario diversity, and wide coverage.
}
\label{tab:compare_dataset}
\vspace{-5pt}
\setlength\tabcolsep{3pt} 
\resizebox{\linewidth}{!}{  
\begin{tabular}{@{}lllcccccc@{}}
\toprule[1.5pt]
\textbf{Benchmark} &\textbf{Anno.} \quad & \textbf{Fire} & \textbf{Smoke} & \textbf{\# Classes} & \textbf{\# Samples} \\
\midrule

BowFire (2015)  & \includegraphics[height=0.9em]{icons8-category-96.pdf} \quad& \textcolor{darkgreen}{\ding{51}} & \textcolor{darkred}{\ding{55}} & 2 & 466  \\

F. Yuan (2015)  & \includegraphics[height=0.9em]{icons8-category-96.pdf} \quad & \textcolor{darkred}{\ding{55}} & \textcolor{darkgreen}{\ding{51}} & 2 & 21,329 \\

Khan (2019)  & \includegraphics[height=0.9em]{icons8-category-96.pdf} \quad & \textcolor{darkred}{\ding{55}} & \textcolor{darkgreen}{\ding{51}} & 4 & 72,012 \\

FD (2020)  & \includegraphics[height=0.9em]{icons8-category-96.pdf} \quad& \textcolor{darkgreen}{\ding{51}} & \textcolor{darkred}{\ding{55}} & 2 & 50,000  \\

L. He (2021) & \includegraphics[height=0.9em]{icons8-category-96.pdf} \quad & \textcolor{darkred}{\ding{55}} & \textcolor{darkgreen}{\ding{51}} & 4 & 33,666 \\

DFAN (2022)  & \includegraphics[height=0.9em]{icons8-category-96.pdf} \quad& \textcolor{darkgreen}{\ding{51}} & \textcolor{darkred}{\ding{55}} & 12 & 3,084  \\

EdgeFireSmoke (2022) & \includegraphics[height=0.9em]{icons8-category-96.pdf}\quad& \textcolor{darkgreen}{\ding{51}} & \textcolor{darkgreen}{\ding{51}} & 8 &  86,775   \\

UFS-data (2022)  & \includegraphics[height=0.9em]{icons8-category-96.pdf} \quad \quad & \textcolor{darkgreen}{\ding{51}} & \textcolor{darkgreen}{\ding{51}} & 8 & 
849,640  \\

MS-FSDB (2024)  &\includegraphics[height=0.9em]{icons8-category-96.pdf} \quad & \textcolor{darkgreen}{\ding{51}} & \textcolor{darkgreen}{\ding{51}} & 35 & 12,586  \\

MAFire-Net (2025)  & \includegraphics[height=0.9em]{icons8-category-96.pdf} \quad \quad & \textcolor{darkgreen}{\ding{51}} & \textcolor{darkred}{\ding{55}}  & 2 & 18,881 \\

\midrule

FiSmo (2017)  & \includegraphics[height=0.9em]{icons8-detective-96.pdf} \includegraphics[height=0.9em]{icons8-top-wide-sidebar-followed-by-partition-at-bottom-96.pdf} \quad& \textcolor{darkgreen}{\ding{51}} & \textcolor{darkgreen}{\ding{51}} & 4 & 9,448  \\

Nemo (2022) & \includegraphics[height=0.9em]{icons8-detective-96.pdf} \includegraphics[height=0.9em]{icons8-category-96.pdf} \quad&  \textcolor{darkred}{\ding{55}} & \textcolor{darkgreen}{\ding{51}} & 4 &  6,702  \\

DFS (2023) & \includegraphics[height=0.9em]{icons8-detective-96.pdf} \includegraphics[height=0.9em]{icons8-category-96.pdf} \quad& \textcolor{darkgreen}{\ding{51}} & \textcolor{darkgreen}{\ding{51}} & 3 & 9,462  \\

HQFSD (2023) &  \includegraphics[height=0.9em]{icons8-detective-96.pdf} \includegraphics[height=0.9em]{icons8-category-96.pdf}\quad& \textcolor{darkgreen}{\ding{51}} & \textcolor{darkgreen}{\ding{51}} & 4 & 12,166  \\

S.Y. Kim (2023) & \includegraphics[height=0.9em]{icons8-detective-96.pdf} \includegraphics[height=0.9em]{icons8-category-96.pdf} \quad&  \textcolor{darkred}{\ding{55}} & \textcolor{darkgreen}{\ding{51}} & 2 &  32,500  \\

DetectiumFire (2026)  & \includegraphics[height=0.9em]{icons8-detective-96.pdf} \includegraphics[height=0.9em]{icons8-category-96.pdf} \includegraphics[height=0.9em]{icons8-chat-left-text-96.pdf}\quad & \textcolor{darkgreen}{\ding{51}} & \textcolor{darkgreen}{\ding{51}}  & 19 & 22,500 \\

SmokeBench$_{\mathrm{(WACV)}}$ (2026) & \includegraphics[height=0.9em]{icons8-detective-96.pdf} \includegraphics[height=0.9em]{icons8-category-96.pdf} \includegraphics[height=0.9em]{icons8-chat-left-text-96.pdf}\quad & \textcolor{darkred}{\ding{55}} & \textcolor{darkgreen}{\ding{51}}  & 2 & 6,046 \\

\midrule

Mlich (2020) & \includegraphics[height=0.9em]{icons8-top-wide-sidebar-followed-by-partition-at-bottom-96.pdf}\quad & \textcolor{darkgreen}{\ding{51}} & \textcolor{darkred}{\ding{55}} & 1 & 6,386  \\

WildfireSpread(2023)  & \includegraphics[height=0.9em]{icons8-top-wide-sidebar-followed-by-partition-at-bottom-96.pdf} \includegraphics[height=0.9em]{icons8-sensor-96-1.pdf}\quad & \textcolor{darkgreen}{\ding{51}} & \textcolor{darkred}{\ding{55}} & 2 & 13,607  \\

MultiFire20K (2024) &  \includegraphics[height=0.9em]{icons8-top-wide-sidebar-followed-by-partition-at-bottom-96.pdf}  \includegraphics[height=0.9em]{icons8-category-96.pdf}\quad& \textcolor{darkgreen}{\ding{51}} & \textcolor{darkgreen}{\ding{51}} & 4 & 20,500 \\

UniInd-FireSmoke(2025)  & \includegraphics[height=0.9em]{icons8-top-wide-sidebar-followed-by-partition-at-bottom-96.pdf} \includegraphics[height=0.9em]{icons8-detective-96.pdf} \quad & \textcolor{darkgreen}{\ding{51}} & \textcolor{darkgreen}{\ding{51}}  & 6 & 30,101  \\

FireSentry (2026)  & \includegraphics[height=0.9em]{icons8-top-wide-sidebar-followed-by-partition-at-bottom-96.pdf} \includegraphics[height=0.9em]{icons8-sensor-96-1.pdf}\quad & \textcolor{darkgreen}{\ding{51}} & \textcolor{darkgreen}{\ding{51}}  & 2 & 20,710 \\

SmokeBench$_{\mathrm{(MM)}}$ (2026)  & Paired Imgs \quad &  \textcolor{darkred}{\ding{55}} & \textcolor{darkgreen}{\ding{51}}  & 43 & 9,975 \\

\midrule
\textbf{SAFIRE (Ours)} &  \includegraphics[height=0.9em]{icons8-chat-left-text-96.pdf} \includegraphics[height=0.9em]{icons8-category-96.pdf} \quad& \textcolor{darkgreen}{\ding{51}} & \textcolor{darkgreen}{\ding{51}} & 20 & 83K/9.7K/193K\\
\bottomrule[1.5pt]
\end{tabular}
}
\vspace{-5pt}
\end{table}

\noindent\textbf{Landscape summary.}\quad
Table~\ref{tab:compare_dataset} summarizes 23 representative fire-smoke datasets (2015-2026). Classification tasks dominate (17/23), with fewer detection or segmentation resources. Only (2/23) include image-caption for evaluating reasoning, and none of the benchmarks include VQA pairs. Besides, over half (13/23) cover only one modality (fire or smoke), despite their frequent co-occurrence. Only a few—\textit{MS-FSDB}~\cite{MS-FSDB}, \textit{DFAN}~\cite{DFAN}, \textit{Detectium}~\cite{liu2026detectiumfire}, and \textit{SmokeBench$_{\mathrm{(MM)}}$}~\cite{jin2025smokebench_mm}—offer relatively diverse scene coverage. 

\noindent\textbf{Rise of multimodal models.}\quad
Recent advances in vision-language modeling, from encoder-only \textit{CLIP}~\cite{CLIP} to multimodal LLMs such as \textit{GLM-4.1V}~\cite{hong2025glm}, \textit{Qwen-3.5}~\cite{qwen3.5}, and \textit{Gemma-3}~\cite{gemma_2025}, enable zero-/few-shot transfer and contextual reasoning. However, prior benchmarks seldom touch the systematical evaluation of fire-smoke reasoning. \textbf{SAFIRE} addresses this gap by introducing diverse real-world scenarios, image-caption pairs, and MCVQA annotations that jointly assess perception, reasoning, and alignment within a unified multimodal framework.


\begin{table}[!t]
    \centering
    \caption{\textbf{SAFIRE dataset statistics across 20 fire-smoke relevant scenarios.} QA-I indicates the image subset used for generating questions (i.e. MCVQA-active images), and QA-P represents the total number of generated QA pairs.}
    \label{tab:image_qna_counts}
    \resizebox{0.9\linewidth}{!}{%
    \begin{tabular}{l r r r}
        \toprule
        \textbf{Scenario} & \textbf{Full Set} & \textbf{QA-I (\%)} & \textbf{QA-P} \\
        
        \multicolumn{4}{l}{\textbf{Accidental Incidents}} \\
        Residential Fire & 4{,}280      & 469 (11.0\%) & 9{,}316 \\
        Explosion                 & 3{,}924      & 463 (11.8\%) & 9{,}257 \\
        Vehicle Fire              & 1{,}616      & 175 (10.8\%) & 3{,}500 \\
        \midrule

        \multicolumn{4}{l}{\textbf{Industrial Operations}} \\
        Aerospace Operation          & 3{,}839      & 454 (11.8\%) & 9{,}056 \\
        Metal Forging             & 2{,}963      & 301 (10.2\%) & 6{,}006 \\
        Flare Stack               & 2{,}298      & 440 (19.1\%) & 8{,}799 \\
        \midrule
        
        \multicolumn{4}{l}{\textbf{Natural Phenomena}} \\
        Volcano                   & 5{,}609      & 728 (13.0\%) & 14{,}561 \\
        Forest Fire              & 5{,}151      & 596 (11.6\%) & 11{,}822 \\
        Open Grassland            & 4{,}912      & 552 (11.2\%) & 11{,}013 \\
        Meteor                    & 2{,}636      & 259 (09.8\%) & 5{,}115 \\
        \midrule

        \multicolumn{4}{l}{\textbf{Recreational Activities}} \\
        Barbecue                  & 6{,}187      & 770 (12.4\%) & 15{,}400 \\
        Firework                  & 6{,}105      & 749 (12.3\%) & 14{,}945 \\
        Campfire-Bonfire          & 5{,}560      & 702 (12.6\%) & 14{,}040 \\
        Torch Parade              & 2{,}922      & 300 (10.3\%) & 5{,}998 \\
        Sky Lantern-Hot Balloon   & 1{,}510      & 194 (12.8\%) & 3{,}880 \\
        \midrule
        \multicolumn{4}{l}{\textbf{Civil Controlled Scenes}} \\
        Waste Disposal            & 5{,}302      & 658 (12.4\%) & 13{,}159 \\
        Gas Stove                 & 5{,}015      & 568 (11.3\%) & 11{,}321 \\
        Incense Burning           & 4{,}970      & 511 (10.3\%) & 10{,}212 \\
        Candlelight               & 4{,}043      & 431 (10.7\%) & 8{,}620 \\
        Smoking                   & 4{,}000      & 348 (08.7\%) & 6{,}884 \\
       
        \midrule
        \textbf{Total} & \textbf{82{,}842} & \textbf{9{,}668 (11.7\%)} & \textbf{192{,}904} \\
        \bottomrule
    \end{tabular}
    } 
    \vspace{-0.5em}
\end{table}

\section{SAFIRE Benchmark}

\begin{figure*}[t]
    \centering
    \includegraphics[width=0.95\linewidth]{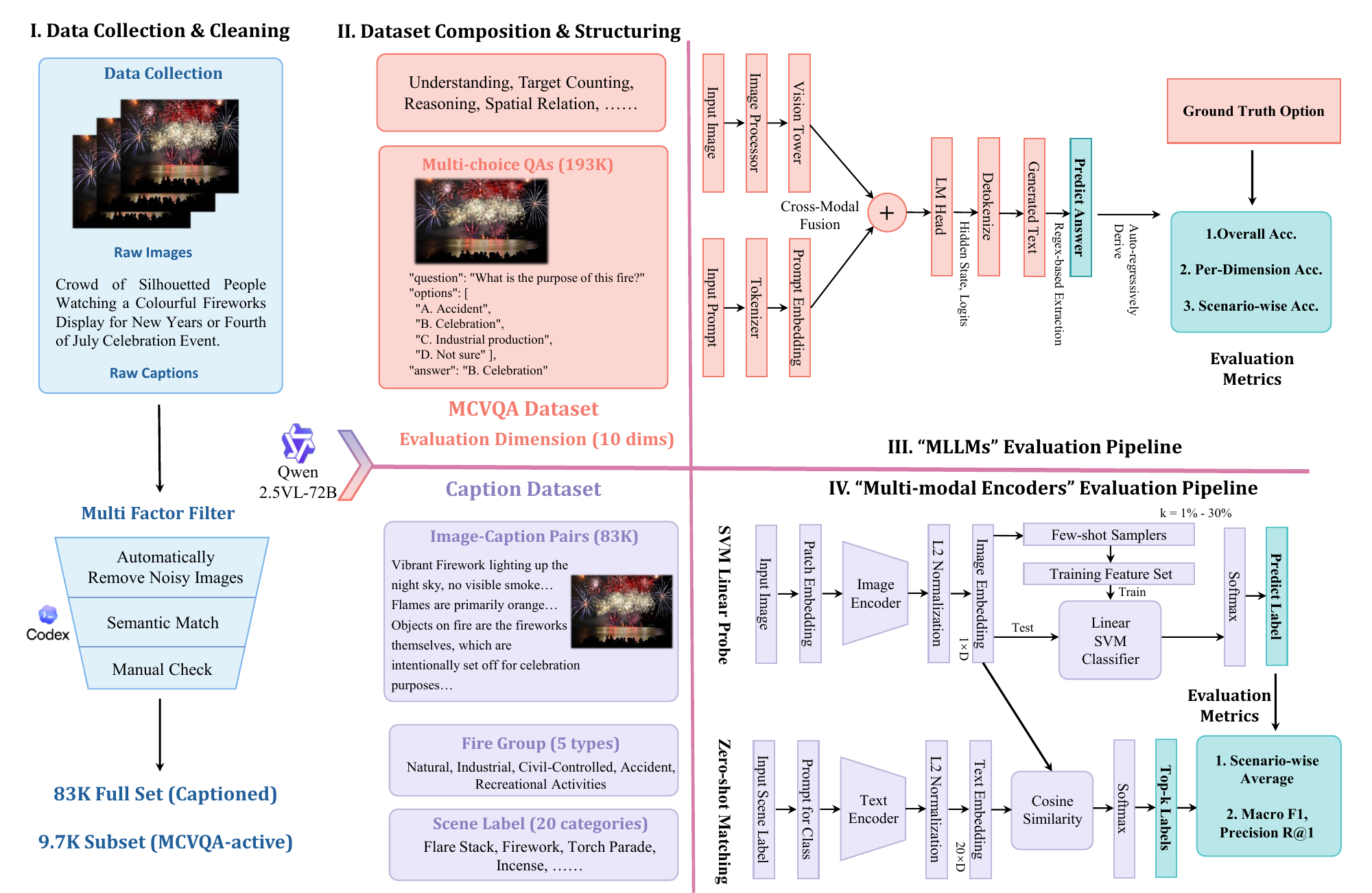}
    \caption{\textbf{SAFIRE pipeline.} Four stages: (I) multi-source collection and filtering; (II) construction of MCVQA and caption branches with hierarchical labels; (III) MLLM evaluation across ten reasoning dimensions; (IV) benchmarking of vision-language encoders under zero-shot and few-shot settings.}
    \label{fig:uni_firesmoke_pipeline}
\end{figure*}

\begin{figure*}[t]
    \centering
    \includegraphics[width=\textwidth]{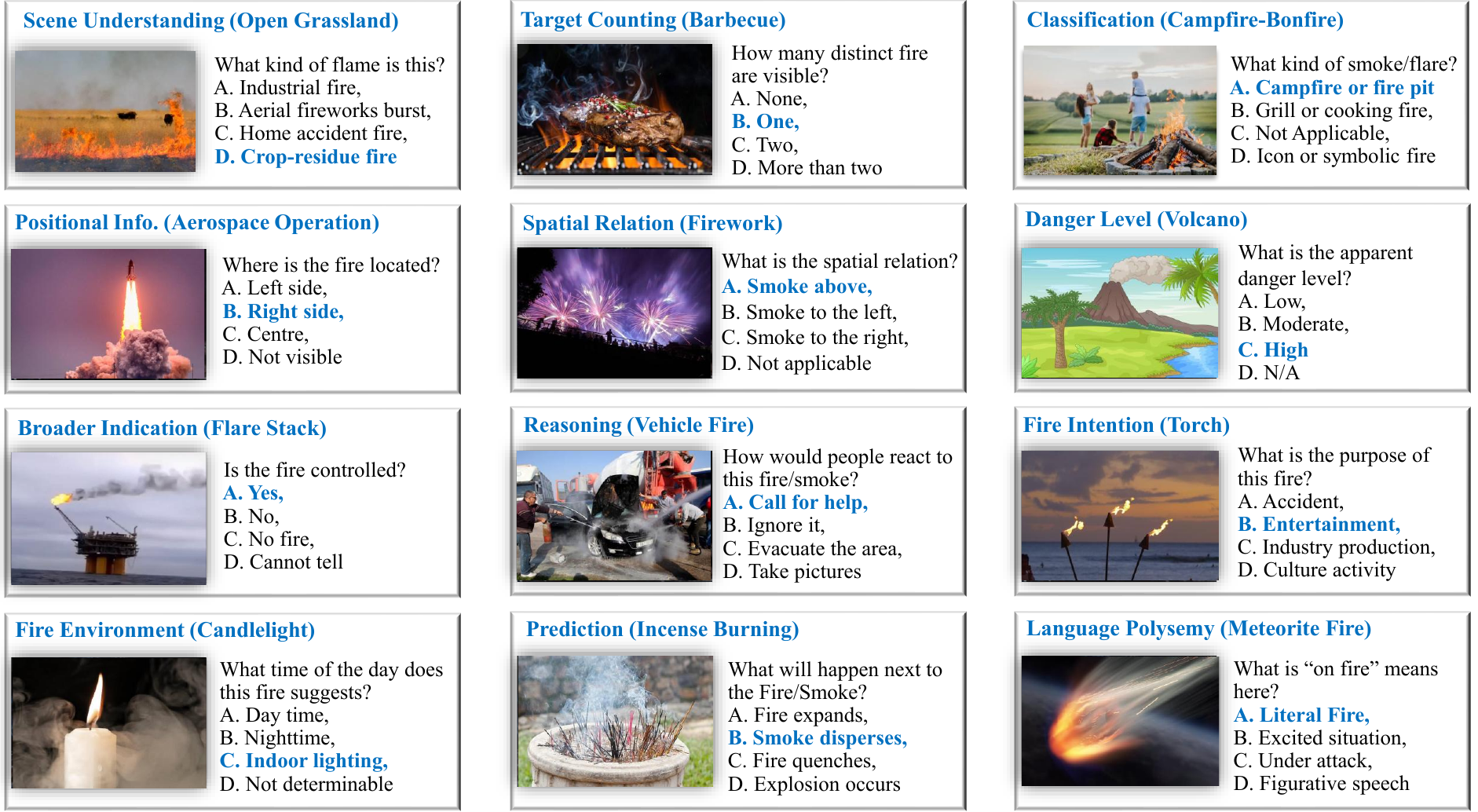}
    \caption{\textbf{Illustrative MCVQA samples.} Each question type (for example, scene understanding or target counting) is aligned with images from distinct fire-smoke scenarios.}
    \label{15examples}
\end{figure*}

We introduce \textbf{SAFIRE}, a large-scale benchmark for evaluating context-aware fire-smoke understanding in MLLMs. It contains 83K captioned-images across 20 real-world scenarios and a curated subset of 9.7K images with 193K multiple-choice VQA (MCVQA) pairs. Complementary image-caption annotations enable both perceptual understanding and higher-level reasoning. Figure~\ref{fig:uni_firesmoke_pipeline} outlines the SAFIRE pipeline, which integrates data collection, dataset structuring, and multimodal evaluation into a unified testbed for fine-grained fire-safety reasoning and zero-shot assessment.


\subsection{Data Collection and Structuring.}
Images and captions are gathered from multiple sources and filtered through a multi-factor filtering (MFF) process combining automatic keyword matching, GPT-5.4-based semantic validation, and manual review (see Appendix \ref{app:dataCleaning}). The pipeline yields 83K high-quality multi-style samples, including both fire-present (positive) and minor fire-absent (negative) images across multiple scenarios, which helps suppress model overconfidence and habitual false-positive fire perception (Fig.~\ref{fig:uni_firesmoke_pipeline}, Stage~I). Subsequently, 9.7K samples are selected to construct two subsets under a unified taxonomy: (i) an MCVQA set spanning ten reasoning dimensions, and (ii) an enhanced image set covering five groups and twenty scene labels (Fig.~\ref{fig:uni_firesmoke_pipeline}, Stage~II).

\textbf{\textit{Annotation Pipeline.}}
All annotations are generated using a structured prompting framework built on Qwen2.5VL-72B-Instruct~\cite{bai2025qwen25vltechnicalreport}. Standardized prompts define the expected outputs across the ten reasoning dimensions to ensure consistent question-answer generation. To reduce hallucinations, the model is instructed to “strictly follow the image content,” and a “None of the above” option is included for ambiguous cases.

\textbf{\textit{MCVQA Quality Control.}}
Annotation quality is ensured through a multi-stage validation pipeline. First, we design scenario-specific prompts to explicitly inform the model of potential contextual biases (e.g., smoke dominance in Incense-burning scenes, or fire dominance in Firework and Metal Forging scenarios). All images, together with their generated MCVQAs, are then passed to GPT-5.4 for verification and correction. Subsequently, we employ majority voting across multiple MLLMs (Fig.~\ref{fig:vqa-clean-pipe}) to identify inconsistent or ambiguous samples, which are further manually reviewed and corrected. This combination of automated validation and human verification provides reliable quality control. Finally, a human evaluation on 2,800 MCVQAs demonstrates an overall annotation accuracy of 87.1\%, with scenario-wise accuracy ranging from 80\% to 96\%. More details about the quality control process, residual error types, and the effect of errors on model performance can be found in Appendix \ref{app:quality control} and \ref{app:annotation_error_analysis} .



\textbf{\textit{Visualization.}} Fig.~\ref{15examples} illustrates representative MCVQA samples, where each reasoning dimension is paired with images from distinct fire-smoke scenarios, highlighting SAFIRE’s diversity and multimodal coverage. More details on dataset collection, cleaning, annotation, prompt design, and structuring are presented in Appendix~\ref{app:dataCleaning} and \ref{app:dataComposition}.

\subsection{MLLM Evaluation.}
We evaluate MLLMs as constrained generators  (Fig.~\ref{fig:uni_firesmoke_pipeline}, Stage~III). Each sample includes one image, its scenario label, and a multiple-choice question with four options (A-D). The prompt instructs the model to respond with a single option letter, which is extracted and compared with the reference label. Questions test skills such as spatial awareness, object counting, intensity estimation, and semantic classification. Accuracy is reported along two axes: per-scenario (robustness across scenes) and per-dimension (skill-level performance). The detailed procedure is described in Appendix~\ref{app:eval_pipeline}.

\subsection{Vision-Language Encoder Evaluation.}
We evaluate underlying capabilities of vision-language encoders (e.g., CLIP~\cite{CLIP}) for fire-smoke scene understanding using both zero-shot and few-shot protocols (Fig.~\ref{fig:uni_firesmoke_pipeline} and Stage IV).

\textbf{\textit{Zero-shot.}}
Each class name (e.g., \textit{campfire}) is inserted into an ensemble of prompt templates such as “A (\textit{photo} or \textit{image} or \textit{scene}) of a campfire.” We compute cosine similarity between frozen image and text features and select the label with the highest score.

\textbf{\textit{SVM Linear Probe.}}
A linear SVM is trained on frozen image embeddings using supervision ratios ranging from 1\% to 30\% with multiple random seeds for sampling, and evaluated on held-out samples. This setting examines how well the learned representations separate fire-related concepts under limited labeled data. Details of data splits, metrics, and training ratios are provided in Appendix~\ref{app:eval_pipeline}.

\section{Experiments}

\begin{table*}[t]
  \centering
  \caption{\textbf{Accuracy (\%) of MLLMs on the 20-scenario fire-smoke MCVQA benchmark.} \textbf{Bold} and \underline{underlined} indicate the highest and second-highest accuracy in each \textbf{column}, and $^{\dagger}$ denotes results from the thinking-enabled version of the model. Avg. is the overall average, Scene Avg. is the macro average.}
  \label{tab:qna_results}
   
\resizebox{\textwidth}{!}{%
\begin{tabular}{l c *{10}{c}}
  \toprule
  \textbf{Model} & \textbf{Size}
    & \multicolumn{4}{c}{\textbf{Natural Phenomena}}
    & \multicolumn{3}{c}{\textbf{Industrial Operations}}
    & \multicolumn{3}{c}{\textbf{Accidental Incidents}} \\
  \cmidrule(lr){3-6}\cmidrule(lr){7-9}\cmidrule(lr){10-12}
    & \textbf{(B)}
    & Grassland & Volcano & Forest & Meteor
    & Aerospace & Flare Stack & Forging
    & Residential & Explos. & Vehicle \\
  \midrule

InternVL3.5         & 8  & 54.13 & 56.96 & 58.85 & 50.60 & 53.13 & 60.85 & 55.48 & 58.35 & 53.06 & 59.46 \\

GLM4.1V  & 9 & 59.58 & 57.85 & 60.13 & 57.66 & 55.56 & \textbf{75.37} & 57.21 & 60.87 & 56.18 & 57.20 \\

Qwen3.5 & 9 & 68.97 & 58.45 & \textbf{63.97} & \textbf{68.28} & \textbf{62.37} & 73.14 & 60.81 & \textbf{65.17} & \textbf{60.73} & 63.03\\
Qwen3.5$^{\dagger}$ & 9 & 64.30 & 57.50 & 61.43 & 59.09 & 59.46 & 74.50 & 56.06 & 61.52 & 56.58 & 60.66\\

Gemma-3             & 12 & 58.42 & 55.66 & 56.91 & 53.29 & 53.09 & 60.68 & 52.08 & 57.85 & 56.54 & 55.31 \\
\midrule

Gemma-3 & 27 & 58.24 & 55.49 & 60.23 & 53.67 & 52.11 & 66.17 & 54.35 & 57.98 & 54.25 & 58.57 \\

Qwen3-VL & 32 & 65.70 & \textbf{61.67} & 63.36 & 60.04 & 60.02 & \underline{74.94} & \textbf{62.29} & 64.00 & 55.97 & \textbf{65.66} \\

Qwen3-VL$^{\dagger}$ & 32 & \underline{69.26} & 60.29 & \underline{63.75} & 61.41 & 61.55 & 72.32 & 59.77 & 63.77 & \underline{58.55} & 64.57 \\

Qwen3.6 & 35A3 & \textbf{69.32} & \underline{61.60} & 63.15 & \underline{62.17} & \underline{61.65} & 74.29 & 61.24 & \underline{64.17} & 57.16 & \underline{64.69}\\

InternVL3.5 & 38 & 66.39 & 58.64 & 61.82 & 59.41 & 58.04 & 73.30 & \underline{61.42} & 62.19 & 55.61 & 61.77 \\

  \midrule
  \textbf{Scene Avg.} & -- & 63.43 & 58.41 & 61.36 & 58.56 & 57.70 & 70.56 & 58.07 & 61.59 & 56.46 & 61.09 \\
  \bottomrule
\end{tabular}
}


  \vspace{0.6em}

\resizebox{\textwidth}{!}{%
\begin{tabular}{l c *{10}{c} |c}
  \toprule
  \textbf{Model} & \textbf{Size}
    & \multicolumn{5}{c}{\textbf{Recreational Activities}}
    & \multicolumn{5}{c}{\textbf{Civil Controlled Scenes}} 
    & \textbf{All} \\
  \cmidrule(lr){3-7}\cmidrule(lr){8-12}\cmidrule(lr){13-13}
    & \textbf{(B)}
    & Firework & SkyLantern & Barbecue & Campfire & Torch
    & Waste & Stove & Incense & Candle & Smoking
    & \textbf{Avg.} \\
  \midrule

InternVL3.5               & 8  & 54.39 & 62.14 & 55.75 & 65.75 & 70.27 & 60.91 & 67.87 & 46.82 & 65.86 & 60.34 & 58.37 \\
GLM4.1V                  & 9 & 52.09  & 57.85 & 50.26 & 60.08 & 66.85 & 59.63 & 70.51 & 58.13 & 69.39 & 60.68 & 59.70 \\
Qwen3.5                  & 9 & 62.18 & 70.38 & 57.55 & 69.93 & 75.02 & 61.03 & \textbf{77.34} & 57.29 & 70.96 & 64.13 & 64.89\\
Qwen3.5$^{\dagger}$       & 9 & 59.72 & 67.10 & 55.02 & 69.12 & 74.84 & 58.87 & 73.83 & 53.19 & 67.85 & 61.20 & 62.17\\
Gemma-3                & 12 & 56.44 & 56.52 & 48.56 & 59.11 & 69.42 & 58.29 & 62.30 & 50.18 & 56.80 & 50.60 & 56.31 \\

\midrule
Gemma-3 & 27 & 58.25 & 62.53 & 50.42 & 65.01 & 73.57 & 60.04 & 65.20 & 45.12 & 61.82 & 55.06 & 58.10 \\

Qwen3-VL & 32 & \textbf{63.66} & 70.54 & \underline{61.10} & \underline{70.02} & 73.87 & \underline{63.97} & 73.64 & \underline{61.31} & \textbf{74.65} & \textbf{68.26} & 65.35 \\

Qwen3-VL$^{\dagger}$ & 32 & 63.29 & \textbf{73.07} & \textbf{61.23} & 69.48 & \underline{76.44} &\textbf{65.73} & 75.07 & \textbf{62.14} & \underline{73.35} & 66.28 & \textbf{65.67} \\
Qwen3.6 & 35A3 & \underline{63.64} & \underline{71.56} & 60.98 & \textbf{70.51} & \textbf{76.92} & 62.71 & \underline{75.92} & 58.77 & 71.23 & \underline{67.50} & \underline{65.53}
\\
InternVL3.5 & 38 & 54.55 & 68.02 & 57.99 & 68.85 & 73.96 & 63.71 & 70.67 & 54.48 & 73.28 & 59.15 & 62.65 \\

  \midrule
  \textbf{Scene Avg.} & -- & 58.82 & 65.97 & 55.89 & 66.79 & 73.12 & 61.49 & 71.23 & 54.74 & 68.52 & 61.32 & --\\
  \bottomrule
\end{tabular}
}
\end{table*}

\subsection{Experimental Setup}
We evaluate both MLLMs and vision-language encoders on the SAFIRE benchmark, as illustrated in Fig.~\ref{fig:uni_firesmoke_pipeline}. Stage~III covers MLLM inference on the MCVQA task, while Stage~IV evaluates vision-language encoders' performance. All MLLM experiments were conducted on an NVIDIA A100 multi-GPU cluster using the vLLM \cite{kwon2023efficient} inference engine with automatic device mapping and \texttt{bfloat16} precision. For vision-language encoders, we used both pretrained general encoders and supervised CLIP+SVM baselines. Experiments are conducted on a curated 9.7K-image test subset. More details about model configurations can be found in Appendix \ref{app:config}

\subsection{MLLM Evaluation Results}

We evaluate MLLMs in \textbf{Stage~III} of the SAFIRE benchmark to assess their ability to solve multimodal multiple-choice questions (MCVQA). This single-run evaluation measures model performance across ten reasoning dimensions and twenty real-world scenarios (under five processes) using structured QA samples. Metrics include overall accuracy, per-scenario performance, and per-dimension accuracy.


\begin{table*}[t]
    \centering
    \caption{
    \textbf{Accuracy (\%) across 10 evaluation dimensions on SAFIRE.}
    \textbf{Bold} indicates the highest accuracy per \textbf{dimension}, while \underline{underlined} indicates the second-highest accuracy, and $^{\dagger}$ denotes results from the thinking-enabled version of the model. Please go to Fig. \ref{15examples} to find typical questions for each dimension.
    }
    \label{tab:qna_question_dim_14models}
    \renewcommand{\arraystretch}{1.3}
    \setlength{\tabcolsep}{2.1pt}
    \footnotesize
    
    \resizebox{\textwidth}{!}{%
    \begin{tabular}{l| *{10}{c} c}
        \toprule
        \textbf{Model} & 
        \makecell{\textbf{Target}\\\textbf{Counting}} & 
        \makecell{\textbf{Class.}} & 
        \makecell{\textbf{General}\\\textbf{Reasoning}} & 
        \makecell{\textbf{Fire/Smoke}\\\textbf{Intention}} & 
        \makecell{\textbf{Emotional}\\\textbf{Response}} & 
        \makecell{\textbf{Polysemy}} & 
        \makecell{\textbf{Fire/Smoke}\\\textbf{Attributes}} & 
        \makecell{\textbf{Spatial}\\\textbf{Corr.}} & 
        \makecell{\textbf{Position}\\\textbf{Identification}} & 
        \makecell{\textbf{Human}\\\textbf{Presence}}\\
        \midrule
        
        InternVL3.5 (8B)        & 67.40 & 73.12 & 49.84 & 67.75 & 54.73 & 59.56 & 58.95 & 34.18 & 55.63 & 88.20  \\
        GLM4.1V (9B)           & \underline{70.12} & 74.52 & 48.87 & 68.32 & 54.24 & 60.60 & 63.02 & 29.38 & 60.95 & \underline{92.85}  \\
        Qwen3.5 (9B)           & 66.41 & 77.29 & 55.81 & 68.50 & \textbf{70.69} & \underline{70.68} & 65.80 & 48.83 & \underline{61.13} & \textbf{93.07} \\
        Qwen3.5$^{\dagger}$ (9B)           & 67.78 & 76.38 & 52.32 & 66.85 & 66.34 & 68.13 & 63.76 & 45.57 & 52.14 & 91.06 \\
        Gemma-3 (12B)           & 42.21 &  71.75 & 49.76 & 71.43 & 45.24 & 66.39 & 56.13 & 38.08 & 54.98 &85.09  \\
        \midrule
        Gemma-3 (27B)           & 48.87 & 76.05 & 50.66 & 69.82 & 49.02 & 63.06 & 57.73 & 42.48 & 56.58 & 87.71 \\

        Qwen3-VL (32B)          & 67.34 & \underline{79.20} & 55.89 & \textbf{73.76} & 69.63 & \textbf{71.61} & 65.98 & 48.23 & 60.59 & 92.15  \\
        Qwen3-VL$^{\dagger}$ (32B)   & 66.78 & \textbf{81.85} & \underline{57.58} & 69.64 & 69.79 & 64.54 & \textbf{66.54} & \textbf{53.77} & 58.50 & 90.49  \\
        Qwen3.6 (35B/A3B)  & \textbf{70.70} & 76.30 & \textbf{57.87} & 70.82 & \underline{69.90} & 65.04 & \underline{66.38} & \underline{49.59} & \textbf{61.16} & 91.62  \\

        InternVL3.5 (38B)        & 63.49 & 77.40 & 54.38 & \underline{71.93} & 67.39 & 66.51 & 63.76 & 38.90 & 54.32 & 90.58  \\

        \midrule
        \textbf{Avg.(\%) } & 63.11 & 76.39 & 53.30 & 69.88 & 61.70 & 65.61 & 62.80 & 42.90 & 57.60 & 90.28 \\
        \bottomrule
    \end{tabular}%
    }
\end{table*}

\noindent\textbf{Performance Across Scenarios.}\quad
Results on the twenty-scenario MCVQA benchmark (Table~\ref{tab:qna_results}) show that most state-of-the-art vision-language models achieve modest performance, with an average accuracy of 61.87\%. This indicates that considerable room remains for improvement in multimodal reasoning under safety-critical conditions. Increasing model size generally improves accuracy, but gains diminish beyond a certain scale; for example, InternVL~38B outperforms its 8B counterpart by around four points, suggesting that scaling alone is insufficient without domain adaptation. Performance also varies widely by scenario: natural phenomena settings (e.g., volcano, meteor) yield lower accuracies, whereas many controlled or familiar scenes (e.g., gas stove, candlelight) perform significantly better. This disparity likely stems from both data imbalance and limited exposure to rare phenomena imagery. Overall, SAFIRE reveals critical weaknesses in current MLLMs and offers guidance for developing models that generalize beyond everyday visual contexts.


\noindent\textbf{Performance Across Reasoning Dimensions.}\quad
Table~\ref{tab:qna_question_dim_14models} compares accuracies across ten evaluation dimensions. Performance varies substantially by task type. Structural dimensions such as \textit{Classification} and \textit{Human Presence} reach higher accuracies (76.4\% and 90.3\%), showing that MLLMs handle basic perceptual tasks well. In contrast, tasks like \textit{Emotional Response} (61.7\%), \textit{Spatial Correlation} (42.9\%), and \textit{General Reasoning} (53.3\%) remain challenging, highlighting persistent difficulty in understanding causality and semantics. Position Identification also lags behind, which is only 57.6\% in average, indicating that models struggle to localize fires under occlusion or subtle conditions. Interestingly, Qwen3-VL$^{\dagger}$ (32B) achieves higher accuracies than its succeeding and larger version (Qwen3.6 35B/A3B) across a few dimensions. Besides, tasks involving semantic ambiguity, such as \textit{Polysemy}, show the widest gap (59.6\%-71.6\%), confirming that context disambiguation remains a major weakness. These results demonstrate SAFIRE’s ability to reveal nuanced strengths and weaknesses that aggregate metrics obscure, emphasizing the value of dimension-wise evaluation for guiding targeted model improvement.




\noindent\textbf{Effect of Reasoning-Enhanced Models.}\quad
We further evaluate reasoning-enhanced (“thinking”) models to enable a direct comparison with standard foundation models. As shown in Table~\ref{tab:qna_results}, even the latest reasoning-oriented architectures achieve only around 62--66\% overall accuracy, indicating that SAFIRE remains a challenging and discriminative benchmark. It effectively captures both performance gaps across models and the incremental gains brought by enhanced reasoning, particularly in fine-grained, safety-critical fire--smoke scenarios. Overall, explicit reasoning yields only modest gains and can even lead to performance degradation. The 9B thinking model is degraded from the corresponding foundation model by 2.7 percentage points, while the 32B series takes a 0.3-percentage-point gain. These results suggest that, despite advances in reasoning capabilities, current MLLMs continue to face significant challenges in achieving robust multimodal reasoning in safety-critical fire--smoke environments.

\begin{figure*}[t]
    \centering
    \includegraphics[width=15.9cm]{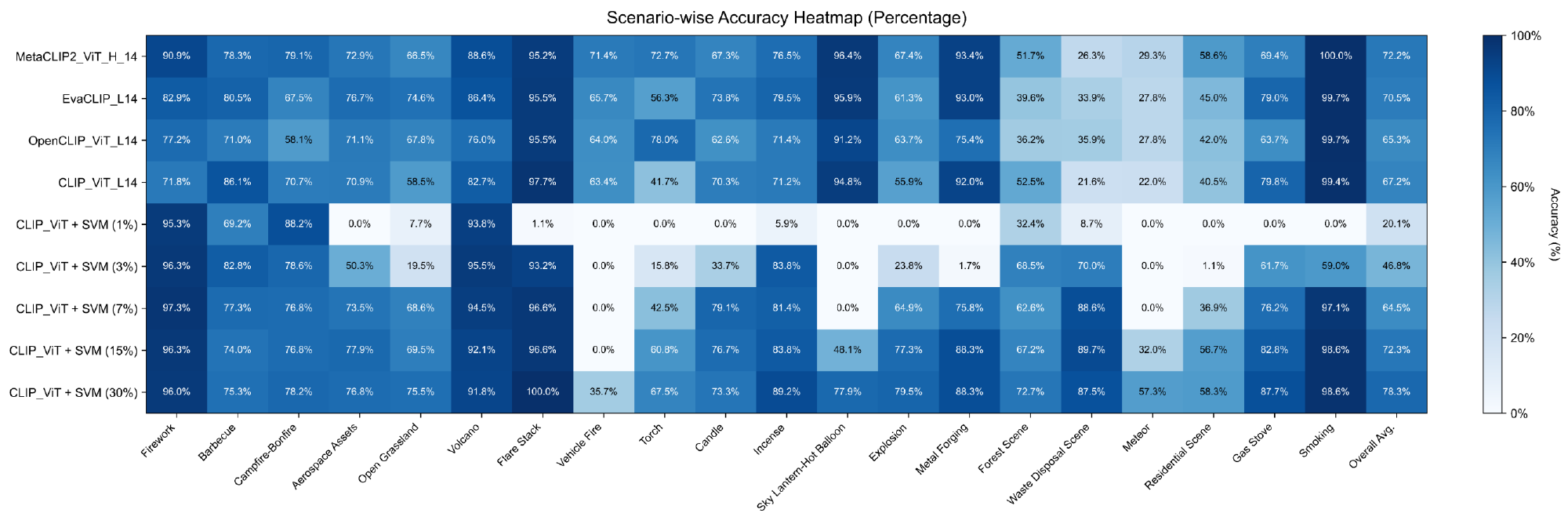}
    \caption{\textbf{Zero-shot accuracy of four multi-modal encoders} (i.e., CLIP-based models) across 20 scenarios in the SAFIRE benchmark. \textbf{The last five rows show results for CLIP+SVM}, which adds a simple tunable classification head. A random guess across 20 classes yields 5\% accuracy.}
    \label{fig:clip_confi}
\end{figure*}

\begin{figure}[t]
    \centering
    \includegraphics[width=\columnwidth]{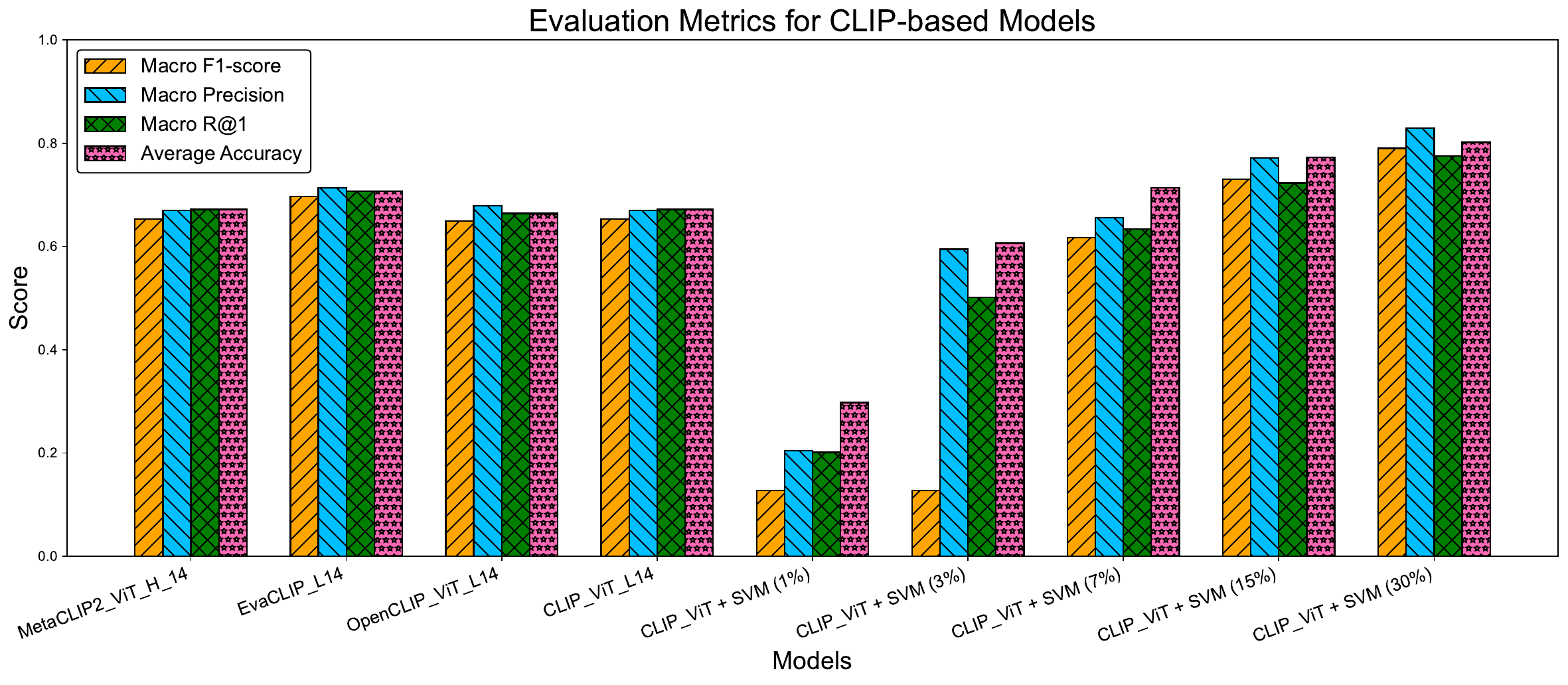}
    \caption{\textbf{Macro performance metrics (F1, Precision, Recall@1), and average accuracy across nine CLIP-based models.} The first four are zero-shot variants, while the latter are CLIP+SVM models with increasing label supervision.}
    \label{fig:clip_histo}
\end{figure}


\noindent\textbf{Evaluation Robustness and Implications.}\quad

Across scenarios and reasoning dimensions, performance gaps among models reflect the combined effects of model scale, architectural design, reasoning capability, and training data quality. Although the evaluated Qwen-family models belong to the same model family as the annotation generator and achieve relatively stronger performance, our expanded generator-family analysis shows that this effect is limited compared with the broader human--model gap; details are provided in Appendix~\ref{app:qwen_family_bias}. In addition, we use the manually audited 2,800-MCVQA subset as a higher-confidence robustness check. The subset is not intended to isolate the causal effect of label correction, since it may differ from the full benchmark in composition and difficulty. Instead, it allows us to examine whether relative model trends remain broadly consistent under cleaner labels. Experimental results suggest that residual annotation noise does not change the main findings. We further include a scenario-prior ablation study in Appendix~\ref{app:scenario_prior_ablation}, demonstrating that the evaluation results are mainly driven by image-grounded reasoning rather than by models exploiting scenario information as shortcuts. A supervised fine-tuning study is also provided in Appendix~\ref{app:mllm_sft}, showing that SAFIRE can provide effective training supervision, with Qwen3-SAFIRE-SFT improving over the Qwen3-VL-8B baseline by 7.81 percentage points on a held-out image-level split. Benefiting from its rich annotations and hierarchical structure, SAFIRE enables fine-grained analysis and provides insights for future research directions, such as domain-specific pretraining, targeted data augmentation for rare or ambiguous scenarios, and tighter integration of visual and textual reasoning. Overall, SAFIRE establishes a comprehensive benchmark for developing reliable, context-aware multimodal systems for safety-critical fire and smoke understanding.


\subsection{Multimodal Encoder Evaluation Results}

Stage~IV of the SAFIRE benchmark evaluates CLIP-based models under both zero-shot inference and supervised linear probing (CLIP+SVM). This single-run analysis examines how well vision-language encoders generalize to diverse fire-smoke scenarios and the effectiveness of lightweight domain adaptation over our high-quality data. Experiments span four zero-shot models and five supervised variants across twenty scenarios (Fig.~\ref{fig:clip_confi},~\ref{fig:clip_histo}).

\noindent\textbf{Scenario-level Analysis.}\quad
Fig.~\ref{fig:clip_confi} visualizes scenario-wise accuracies. Zero-shot variants such as \texttt{MetaCLIP2} \cite{chuang2026metaclip} and \texttt{EvaCLIP} \cite{sun2023eva} substantially outperform the 5\% random baseline and reach up to 72.2\% accuracy. Highly salient categories such as \textit{Smoking}, \textit{Flare Stack}, and \textit{Volcano} exceed 80\% accuracy, while visually cluttered or ambiguous scenes such as \textit{Meteor}, \textit{Residential Scene}, and \textit{Waste Disposal Scene} show weaker results. Model-wise variation is also evident: \textit{Firework} achieves over 90\% accuracy in \texttt{MetaCLIP} but only 71\% in \texttt{CLIP\_ViT\_L14}, suggesting reliance on global scene cues.

\noindent\textbf{Metric-level Analysis.}\quad Fig.~\ref{fig:clip_histo} compares macro-level performance across nine encoders. Among zero-shot models, \texttt{EvaCLIP} achieves the best balance (F1=0.70, Recall@1=0.71, Precision=0.71), while other generic models lag behind. Although \texttt{MetaCLIP} attains the highest overall accuracy, its macro-level metrics remain consistently lower than those of \texttt{EvaCLIP}, indicating weaker class-balanced generalization.


Supervised variants (CLIP+SVM) show consistent gains as label supervision increases. With 30\% supervision, CLIP+SVM achieves F1=0.79 and Precision=0.83, while even 7\% supervision improves performance to F1=0.63—approaching most zero-shot baselines. These results highlight the efficiency of lightweight fine-tuning for specialized domains.

\noindent\textbf{Summary and Implications.}\quad
SAFIRE provides a challenging yet structured benchmark for vision-language alignment. Despite the difficulty of the 20-class classification task, strong zero-shot models such as \texttt{MetaCLIP2} reaches 72.2\% accuracy, confirming the dataset’s semantic separability and visual grounding. Overall, SAFIRE’s diversity and granularity make it also a robust testbed for classification and retrieval-style evaluation under both zero- and linear probe settings, emphasizing the potential of lightweight adaptation for improving encoder performance in fire-smoke understanding.


\section{Conclusion}
We presented \textbf{SAFIRE}, a large-scale benchmark for context-aware fire-smoke understanding. The dataset comprises over \textbf{83K}-captioned images across \textbf{20} real-world scenarios and \textbf{193K} multiple-choice VQA pairs, organized under five groups (natural, industrial, accidental, recreational, civil controlled). SAFIRE supports two complementary evaluations: multi-dimensional reasoning with MLLMs via MCVQA, and zero-/few-shot scene classification with vision-language encoders. On SAFIRE, ten open-source MLLMs average \textbf{61.9\%} MCVQA accuracy, while CLIP-style encoders reach up to \textbf{72.2\%} in zero-shot classification; adapting vision encoders with only 7\% supervised data boosts fire-scene classification accuracy from 20.1\% to 64.5\%. These results highlight remaining gaps in safety-critical reasoning and the value of lightweight domain adaptation. We hope SAFIRE serves as a standardized testbed for building more reliable multimodal systems in fire- and smoke-related scenarios.

\section*{Limitations} \label{limit}
While SAFIRE provides a strong foundation for fire and smoke understanding, several aspects offer opportunities for further improvement.

First, retrieval and alignment evaluations rely on limited natural-language prompts (e.g., “A (\textit{photo} or \textit{image} or \textit{scene}) of a {Vehicle fire}”). Exploring more diverse prompts, including syntactic and semantic variations, as well as prompt tuning or renaming strategies (e.g., replacing “waste disposal fire” with “dumpster and garbage fire”), may further improve generalization and text–image alignment in safety-critical scenarios.

Second, the evaluated models are pretrained on general-domain corpora. Our adaptation is limited to the linear probe and one supervised fine-tuning configuration; broader evaluation across architectures and adaptation methods remains future work.

Finally, most annotations are generated by a vision-language model and refined through GPT-5.4 and small-model voting, with human verification for flagged samples. While this pipeline enables scalable and consistent data creation, biases may still persist due to the model's capability at the time of the Benchmark's design. Ongoing auditing (e.g., use different SOTA models to vote or correct question-answer pairs) and community-driven validation are needed to ensure fairness and reliability.

Overall, these limitations point to clear directions for future work, and addressing them will further strengthen SAFIRE as a reliable benchmark for safety-critical multimodal systems.

\section*{Acknowledgements}
This research was funded by Khalifa University of Science and Technology through the Faculty Start-Ups under Project ID: KU-INT-FSU-2005-8474000775.

\bibliography{custom}

@inproceedings{jin2025smokebench_mm,
  title={SmokeBench: A Real-World Dataset for Surveillance Image Desmoking in Early-Stage Fire Scenes},
  author={Jin, Wenzhuo and Yang, Qianfeng and Wu, Xianhao and Chen, Hongming and Li, Pengpeng and Chen, Xiang},
  booktitle={Proceedings of the 33rd ACM International Conference on Multimedia},
  pages={12722--12728},
  year={2025}
}

@article{liu2026detectiumfire,
  title={Detectiumfire: A comprehensive multi-modal dataset bridging vision and language for fire understanding},
  author={Liu, Zixuan and H Khajavi, Siavash and Jiang, Guangkai},
  journal={Advances in Neural Information Processing Systems},
  volume={38},
  year={2026}
}

@inproceedings{qi2026smokebench_wacv,
  title={Smokebench: Evaluating multimodal large language models for wildfire smoke detection},
  author={Qi, Tianye and Li, Weihao and Barnes, Nick},
  booktitle={2026 IEEE/CVF Winter Conference on Applications of Computer Vision (WACV)},
  pages={1043--1053},
  year={2026},
  organization={IEEE}
}

@inproceedings{zhou2026firesentry,
  title={FireSentry: A multi-modal spatio-temporal benchmark dataset for fine-grained wildfire spread forecasting},
  author={Zhou, Nan and Wang, Huandong and Li, Jiahao and Li, Han and Song, Yali and Wang, Qiuhua and Li, Yong and Chen, Xinlei},
  booktitle={Proceedings of the 32nd ACM SIGKDD Conference on Knowledge Discovery and Data Mining V. 1},
  pages={2935--2946},
  year={2026}
}

@misc{qwen3.5,
    title  = {{Qwen3.5}: Towards Native Multimodal Agents},
    author = {{Qwen Team}},
    month  = {February},
    year   = {2026},
    url    = {https://qwen.ai/blog?id=qwen3.5}
}

@misc{qwen36_35b_a3b,
    title = {{Qwen3.6-35B-A3B}: Agentic Coding Power, Now Open to All},
    url = {https://qwen.ai/blog?id=qwen3.6-35b-a3b},
    author = {{Qwen Team}},
    month = {April},
    year = {2026}
}

@inproceedings{chuang2026metaclip,
title={Meta {CLIP} 2: A Worldwide Scaling Recipe},
author={Yung-Sung Chuang and Yang Li and Dong Wang and Ching-Feng Yeh and Kehan Lyu and Ramya Raghavendra and James R. Glass and LIFEI HUANG and Jason E Weston and Luke Zettlemoyer and Xinlei Chen and Zhuang Liu and Saining Xie and Wen-tau Yih and Shang-Wen Li and Hu Xu},
booktitle={The Thirty-ninth Annual Conference on Neural Information Processing Systems},
year={2025},
url={https://openreview.net/forum?id=aYRNINhNGV}
}

@article{smokeseg,
  title={An effective multi-scale interactive fusion network with hybrid Transformer and CNN for smoke image segmentation},
  author={Li, Kang and Yuan, Feiniu and Wang, Chunmei},
  journal={Pattern Recognition},
  volume={159},
  pages={111177},
  year={2025},
  publisher={Elsevier}
}

@article{zhu2025multiscale_det,
  title={Multiscale wildfire and smoke detection in complex drone forest environments based on YOLOv8},
  author={Zhu, Wenyu and Niu, Shanwei and Yue, Jixiang and Zhou, Yangli},
  journal={Scientific Reports},
  volume={15},
  number={1},
  pages={2399},
  year={2025},
  publisher={Nature Publishing Group UK London}
}

@article{du2025firemultiformer,
  title={FireMultiFormer: A Transformer-Based Multimodal Approach for Predicting Arc Fault Fires with Fire Images and Fault Signals},
  author={Du, Liwei and Xu, Zhihong and Gao, Jianhong and Chen, Duanyu},
  journal={IEEE Transactions on Instrumentation and Measurement},
  year={2025},
  publisher={IEEE}
}

@inproceedings{fsdtechnical,
  title={Fire and Smoke Detection with Burning Intensity Representation},
  author={Han, Xiaoyi and Wu, Yanfei and Pu, Nan and Feng, Zunlei and Zhang, Qifei and Bei, Yijun and Cheng, Lechao},
  booktitle={Proceedings of the 6th ACM International Conference on Multimedia in Asia},
  pages={1--8},
  year={2024}
}

@misc{gemma_2025,
      title={Gemma 3 Technical Report}, 
      author={Gemma Team and Aishwarya Kamath and Johan Ferret and Shreya Pathak and Nino Vieillard and Ramona Merhej and Sarah Perrin and Tatiana Matejovicova and Alexandre Ramé and Morgane Rivière and Louis Rouillard and Thomas Mesnard and Geoffrey Cideron and Jean-bastien Grill and Sabela Ramos and Edouard Yvinec and Michelle Casbon and Etienne Pot and Ivo Penchev and Gaël Liu and Francesco Visin and Kathleen Kenealy and Lucas Beyer and Xiaohai Zhai and Anton Tsitsulin and Robert Busa-Fekete and Alex Feng and Noveen Sachdeva and Benjamin Coleman and Yi Gao and Basil Mustafa and Iain Barr and Emilio Parisotto and David Tian and Matan Eyal and Colin Cherry and Jan-Thorsten Peter and Danila Sinopalnikov and Surya Bhupatiraju and Rishabh Agarwal and Mehran Kazemi and Dan Malkin and Ravin Kumar and David Vilar and Idan Brusilovsky and Jiaming Luo and Andreas Steiner and Abe Friesen and Abhanshu Sharma and Abheesht Sharma and Adi Mayrav Gilady and Adrian Goedeckemeyer and Alaa Saade and Alex Feng and Alexander Kolesnikov and Alexei Bendebury and Alvin Abdagic and Amit Vadi and András György and André Susano Pinto and Anil Das and Ankur Bapna and Antoine Miech and Antoine Yang and Antonia Paterson and Ashish Shenoy and Ayan Chakrabarti and Bilal Piot and Bo Wu and Bobak Shahriari and Bryce Petrini and Charlie Chen and Charline Le Lan and Christopher A. Choquette-Choo and CJ Carey and Cormac Brick and Daniel Deutsch and Danielle Eisenbud and Dee Cattle and Derek Cheng and Dimitris Paparas and Divyashree Shivakumar Sreepathihalli and Doug Reid and Dustin Tran and Dustin Zelle and Eric Noland and Erwin Huizenga and Eugene Kharitonov and Frederick Liu and Gagik Amirkhanyan and Glenn Cameron and Hadi Hashemi and Hanna Klimczak-Plucińska and Harman Singh and Harsh Mehta and Harshal Tushar Lehri and Hussein Hazimeh and Ian Ballantyne and Idan Szpektor and Ivan Nardini and Jean Pouget-Abadie and Jetha Chan and Joe Stanton and John Wieting and Jonathan Lai and Jordi Orbay and Joseph Fernandez and Josh Newlan and Ju-yeong Ji and Jyotinder Singh and Kat Black and Kathy Yu and Kevin Hui and Kiran Vodrahalli and Klaus Greff and Linhai Qiu and Marcella Valentine and Marina Coelho and Marvin Ritter and Matt Hoffman and Matthew Watson and Mayank Chaturvedi and Michael Moynihan and Min Ma and Nabila Babar and Natasha Noy and Nathan Byrd and Nick Roy and Nikola Momchev and Nilay Chauhan and Noveen Sachdeva and Oskar Bunyan and Pankil Botarda and Paul Caron and Paul Kishan Rubenstein and Phil Culliton and Philipp Schmid and Pier Giuseppe Sessa and Pingmei Xu and Piotr Stanczyk and Pouya Tafti and Rakesh Shivanna and Renjie Wu and Renke Pan and Reza Rokni and Rob Willoughby and Rohith Vallu and Ryan Mullins and Sammy Jerome and Sara Smoot and Sertan Girgin and Shariq Iqbal and Shashir Reddy and Shruti Sheth and Siim Põder and Sijal Bhatnagar and Sindhu Raghuram Panyam and Sivan Eiger and Susan Zhang and Tianqi Liu and Trevor Yacovone and Tyler Liechty and Uday Kalra and Utku Evci and Vedant Misra and Vincent Roseberry and Vlad Feinberg and Vlad Kolesnikov and Woohyun Han and Woosuk Kwon and Xi Chen and Yinlam Chow and Yuvein Zhu and Zichuan Wei and Zoltan Egyed and Victor Cotruta and Minh Giang and Phoebe Kirk and Anand Rao and Kat Black and Nabila Babar and Jessica Lo and Erica Moreira and Luiz Gustavo Martins and Omar Sanseviero and Lucas Gonzalez and Zach Gleicher and Tris Warkentin and Vahab Mirrokni and Evan Senter and Eli Collins and Joelle Barral and Zoubin Ghahramani and Raia Hadsell and Yossi Matias and D. Sculley and Slav Petrov and Noah Fiedel and Noam Shazeer and Oriol Vinyals and Jeff Dean and Demis Hassabis and Koray Kavukcuoglu and Clement Farabet and Elena Buchatskaya and Jean-Baptiste Alayrac and Rohan Anil and Dmitry and Lepikhin and Sebastian Borgeaud and Olivier Bachem and Armand Joulin and Alek Andreev and Cassidy Hardin and Robert Dadashi and Léonard Hussenot},
      year={2025},
      eprint={2503.19786},
      archivePrefix={arXiv},
      primaryClass={cs.CL},
      url={https://arxiv.org/abs/2503.19786}, 
}

@article{sun2023eva,
  title={Eva-clip: Improved training techniques for clip at scale},
  author={Sun, Quan and Fang, Yuxin and Wu, Ledell and Wang, Xinlong and Cao, Yue},
  journal={arXiv preprint arXiv:2303.15389},
  year={2023}
}

@inproceedings{openclipcvpr,
  title={Reproducible scaling laws for contrastive language-image learning},
  author={Cherti, Mehdi and Beaumont, Romain and Wightman, Ross and Wortsman, Mitchell and Ilharco, Gabriel and Gordon, Cade and Schuhmann, Christoph and Schmidt, Ludwig and Jitsev, Jenia},
  booktitle={Proceedings of the IEEE/CVF Conference on Computer Vision and Pattern Recognition},
  pages={2818--2829},
  year={2023}
}

@article{smoke2022survey,
  title={A survey on vision-based outdoor smoke detection techniques for environmental safety},
  author={Chaturvedi, Shubhangi and Khanna, Pritee and Ojha, Aparajita},
  journal={ISPRS Journal of Photogrammetry and Remote Sensing},
  volume={185},
  pages={158--187},
  year={2022},
  publisher={Elsevier}
}

@article{liu2023visual,
  title={Visual instruction tuning},
  author={Liu, Haotian and Li, Chunyuan and Wu, Qingyang and Lee, Yong Jae},
  journal={Advances in neural information processing systems},
  volume={36},
  pages={34892--34916},
  year={2023}
}

@data{UniInd,
doi = {10.21227/6a33-k522},
url = {https://dx.doi.org/10.21227/6a33-k522},
author = {Pengfei Li and Said Boumaraf and Muaz Al Radi and Sicheng Zhang},
publisher = {IEEE Dataport},
title = {UniInd-FireSmoke: Unified Industrial Fire-Smoke Dataset},
year = {2025} }

@article{yazdi2022nemo,
  title={Nemo: An open-source transformer-supercharged benchmark for fine-grained wildfire smoke detection},
  author={Yazdi, Amirhessam and Qin, Heyang and Jordan, Connor B and Yang, Lei and Yan, Feng},
  journal={Remote Sensing},
  volume={14},
  number={16},
  pages={3979},
  year={2022},
  publisher={MDPI}
}

@inproceedings{MS-FSDB,
  title={Benchmarking Multi-Scene Fire and Smoke Detection},
  author={Han, Xiaoyi and Pu, Nan and Feng, Zunlei and Bei, Yijun and Zhang, Qifei and Cheng, Lechao and Xue, Liang},
  booktitle={Chinese Conference on Pattern Recognition and Computer Vision (PRCV)},
  pages={203--218},
  year={2024},
  organization={Springer}
}

@article{khandataset,
  title={Energy-efficient deep CNN for smoke detection in foggy IoT environment},
  author={Khan, Salman and Muhammad, Khan and Mumtaz, Shahid and Baik, Sung Wook and de Albuquerque, Victor Hugo C},
  journal={IEEE Internet of Things Journal},
  volume={6},
  number={6},
  pages={9237--9245},
  year={2019},
  publisher={IEEE}
}

@article{FD,
  title={An efficient fire detection method based on multiscale feature extraction, implicit deep supervision and channel attention mechanism},
  author={Li, Songbin and Yan, Qiandong and Liu, Peng},
  journal={IEEE Transactions on Image Processing},
  volume={29},
  pages={8467--8475},
  year={2020},
  publisher={IEEE}
}

@article{MAFire,
  title={Enhancing real-time fire detection: An effective multi-attention network and a fire benchmark},
  author={Khan, Taimoor and Khan, Zulfiqar Ahmad and Choi, Chang},
  journal={Neural Computing and Applications},
  volume={37},
  number={18},
  pages={11693--11707},
  year={2025},
  publisher={Springer}
}

@article{DFS,
  title={A dataset for fire and smoke object detection},
  author={Wu, Siyuan and Zhang, Xinrong and Liu, Ruqi and Li, Binhai},
  journal={Multimedia Tools and Applications},
  volume={82},
  number={5},
  pages={6707--6726},
  year={2023},
  publisher={Springer}
}

@inproceedings{hqfsd,
  title={High Quality Fire Smoke Dataset: A Benchmark for Fire and Smoke Detection},
  author={Qian, Jialong and Hong, Chaoqun and Zhang, Kejie and Huang, Jianglong},
  booktitle={Proceedings of the 2nd International Workshop on Multimedia Content Generation and Evaluation: New Methods and Practice},
  pages={26--35},
  year={2024}
}

@article{LHe,
  title={Efficient attention based deep fusion CNN for smoke detection in fog environment},
  author={He, Lijun and Gong, Xiaoli and Zhang, Sirou and Wang, Liejun and Li, Fan},
  journal={Neurocomputing},
  volume={434},
  pages={224--238},
  year={2021},
  publisher={Elsevier}
}

@article{almeida2022edgefiresmoke,
  title={EdgeFireSmoke: A novel lightweight CNN model for real-time video fire--smoke detection},
  author={Almeida, Jefferson Silva and Huang, Chenxi and Nogueira, Fabr{\'\i}cio Gonzalez and Bhatia, Surbhi and de Albuquerque, Victor Hugo C},
  journal={IEEE Transactions on Industrial Informatics},
  volume={18},
  number={11},
  pages={7889--7898},
  year={2022},
  publisher={IEEE}
}

@article{DFAN,
  title={Optimized dual fire attention network and medium-scale fire classification benchmark},
  author={Yar, Hikmat and Hussain, Tanveer and Agarwal, Mohit and Khan, Zulfiqar Ahmad and Gupta, Suneet Kumar and Baik, Sung Wook},
  journal={IEEE Transactions on Image Processing},
  volume={31},
  pages={6331--6343},
  year={2022},
  publisher={IEEE}
}

@inproceedings{chino2015bowfire,
  title={Bowfire: detection of fire in still images by integrating pixel color and texture analysis},
  author={Chino, Daniel YT and Avalhais, Letricia PS and Rodrigues, Jose F and Traina, Agma JM},
  booktitle={2015 28th SIBGRAPI conference on graphics, patterns and images},
  pages={95--102},
  year={2015},
  organization={IEEE}
}

@article{yuan2015real,
  title={Real-time image smoke detection using staircase searching-based dual threshold AdaBoost and dynamic analysis},
  author={Yuan, Feiniu and Fang, Zhijun and Wu, Shiqian and Yang, Yong and Fang, Yuming},
  journal={IET Image Processing},
  volume={9},
  number={10},
  pages={849--856},
  year={2015},
  publisher={IET}
}

@article{gerard2023wildfirespreadts,
  title={Wildfirespreadts: A dataset of multi-modal time series for wildfire spread prediction},
  author={Gerard, Sebastian and Zhao, Yu and Sullivan, Josephine},
  journal={Advances in Neural Information Processing Systems},
  volume={36},
  pages={74515--74529},
  year={2023}
}

@inproceedings{fismo,
  title={Fismo: A compilation of datasets from emergency situations for fire and smoke analysis},
  author={Cazzolato, Mirela T and Avalhais, Letricia and Chino, Daniel and Ramos, Jonathan S and de Souza, Jessica A and Rodrigues-Jr, Jose F and Traina, A},
  booktitle={Brazilian symposium on databases-SBBD},
  pages={213--223},
  year={2017},
  organization={SBC Uberl{\^a}ndia, Brazil}
}

@article{boumaraf2025vision,
  title={Vision-based air-flow monitoring in industrial flares system design using deep convolutional neural networks},
  author={Boumaraf, Said and Al Radi, Muaz and Abdelhafez, Fares Oussama and Li, Pengfei and Al Awadhi, Khalid Yousef and Karki, Hamad and Dhelim, Sahraoui and Werghi, Naoufel},
  journal={Expert Systems with Applications},
  pages={126733},
  year={2025},
  publisher={Elsevier}
}

@article{hosseini2022ufs,
  title={UFS-Net: A unified flame and smoke detection method for early detection of fire in video surveillance applications using CNNs},
  author={Hosseini, Ali and Hashemzadeh, Mahdi and Farajzadeh, Nacer},
  journal={Journal of Computational Science},
  volume={61},
  pages={101638},
  year={2022},
  publisher={Elsevier}
}

@InProceedings{CLIP,
  title = 	 {Learning Transferable Visual Models From Natural Language Supervision},
  author =       {Radford, Alec and Kim, Jong Wook and Hallacy, Chris and Ramesh, Aditya and Goh, Gabriel and Agarwal, Sandhini and Sastry, Girish and Askell, Amanda and Mishkin, Pamela and Clark, Jack and Krueger, Gretchen and Sutskever, Ilya},
  booktitle = 	 {Proceedings of the 38th International Conference on Machine Learning},
  pages = 	 {8748--8763},
  year = 	 {2021},
  editor = 	 {Meila, Marina and Zhang, Tong},
  volume = 	 {139},
  series = 	 {Proceedings of Machine Learning Research},
  month = 	 {18--24 Jul},
  publisher =    {PMLR},
  url = 	 {https://proceedings.mlr.press/v139/radford21a.html}
}

@misc{bai2025qwen25vltechnicalreport,
      title={Qwen2.5-VL Technical Report}, 
      author={Shuai Bai and Keqin Chen and Xuejing Liu and Jialin Wang and Wenbin Ge and Sibo Song and Kai Dang and Peng Wang and Shijie Wang and Jun Tang and Humen Zhong and Yuanzhi Zhu and Mingkun Yang and Zhaohai Li and Jianqiang Wan and Pengfei Wang and Wei Ding and Zheren Fu and Yiheng Xu and Jiabo Ye and Xi Zhang and Tianbao Xie and Zesen Cheng and Hang Zhang and Zhibo Yang and Haiyang Xu and Junyang Lin},
      year={2025},
      eprint={2502.13923},
      archivePrefix={arXiv},
      primaryClass={cs.CV},
      url={https://arxiv.org/abs/2502.13923}, 
}

@inproceedings{kwon2023efficient,
  title={Efficient memory management for large language model serving with pagedattention},
  author={Kwon, Woosuk and Li, Zhuohan and Zhuang, Siyuan and Sheng, Ying and Zheng, Lianmin and Yu, Cody Hao and Gonzalez, Joseph and Zhang, Hao and Stoica, Ion},
  booktitle={Proceedings of the 29th symposium on operating systems principles},
  pages={611--626},
  year={2023}
}

@misc{bai2025qwen3,
      title={Qwen3-VL Technical Report}, 
      author={Shuai Bai and Yuxuan Cai and Ruizhe Chen and Keqin Chen and Xionghui Chen and Zesen Cheng and Lianghao Deng and Wei Ding and Chang Gao and Chunjiang Ge and Wenbin Ge and Zhifang Guo and Qidong Huang and Jie Huang and Fei Huang and Binyuan Hui and Shutong Jiang and Zhaohai Li and Mingsheng Li and Mei Li and Kaixin Li and Zicheng Lin and Junyang Lin and Xuejing Liu and Jiawei Liu and Chenglong Liu and Yang Liu and Dayiheng Liu and Shixuan Liu and Dunjie Lu and Ruilin Luo and Chenxu Lv and Rui Men and Lingchen Meng and Xuancheng Ren and Xingzhang Ren and Sibo Song and Yuchong Sun and Jun Tang and Jianhong Tu and Jianqiang Wan and Peng Wang and Pengfei Wang and Qiuyue Wang and Yuxuan Wang and Tianbao Xie and Yiheng Xu and Haiyang Xu and Jin Xu and Zhibo Yang and Mingkun Yang and Jianxin Yang and An Yang and Bowen Yu and Fei Zhang and Hang Zhang and Xi Zhang and Bo Zheng and Humen Zhong and Jingren Zhou and Fan Zhou and Jing Zhou and Yuanzhi Zhu and Ke Zhu},
      year={2025},
      eprint={2511.21631},
      archivePrefix={arXiv},
      primaryClass={cs.CV},
      url={https://arxiv.org/abs/2511.21631}, 
}

@misc{wang2025internvl3_5,
      title={InternVL3.5: Advancing Open-Source Multimodal Models in Versatility, Reasoning, and Efficiency}, 
      author={Weiyun Wang and Zhangwei Gao and Lixin Gu and Hengjun Pu and Long Cui and Xingguang Wei and Zhaoyang Liu and Linglin Jing and Shenglong Ye and Jie Shao and Zhaokai Wang and Zhe Chen and Hongjie Zhang and Ganlin Yang and Haomin Wang and Qi Wei and Jinhui Yin and Wenhao Li and Erfei Cui and Guanzhou Chen and Zichen Ding and Changyao Tian and Zhenyu Wu and Jingjing Xie and Zehao Li and Bowen Yang and Yuchen Duan and Xuehui Wang and Zhi Hou and Haoran Hao and Tianyi Zhang and Songze Li and Xiangyu Zhao and Haodong Duan and Nianchen Deng and Bin Fu and Yinan He and Yi Wang and Conghui He and Botian Shi and Junjun He and Yingtong Xiong and Han Lv and Lijun Wu and Wenqi Shao and Kaipeng Zhang and Huipeng Deng and Biqing Qi and Jiaye Ge and Qipeng Guo and Wenwei Zhang and Songyang Zhang and Maosong Cao and Junyao Lin and Kexian Tang and Jianfei Gao and Haian Huang and Yuzhe Gu and Chengqi Lyu and Huanze Tang and Rui Wang and Haijun Lv and Wanli Ouyang and Limin Wang and Min Dou and Xizhou Zhu and Tong Lu and Dahua Lin and Jifeng Dai and Weijie Su and Bowen Zhou and Kai Chen and Yu Qiao and Wenhai Wang and Gen Luo},
      year={2025},
      eprint={2508.18265},
      archivePrefix={arXiv},
      primaryClass={cs.CV},
      url={https://arxiv.org/abs/2508.18265}, 
}

@misc{hong2025glm,
      title={GLM-4.5V and GLM-4.1V-Thinking: Towards Versatile Multimodal Reasoning with Scalable Reinforcement Learning}, 
      author={V Team and Wenyi Hong and Wenmeng Yu and Xiaotao Gu and Guo Wang and Guobing Gan and Haomiao Tang and Jiale Cheng and Ji Qi and Junhui Ji and Lihang Pan and Shuaiqi Duan and Weihan Wang and Yan Wang and Yean Cheng and Zehai He and Zhe Su and Zhen Yang and Ziyang Pan and Aohan Zeng and Baoxu Wang and Bin Chen and Boyan Shi and Changyu Pang and Chenhui Zhang and Da Yin and Fan Yang and Guoqing Chen and Haochen Li and Jiale Zhu and Jiali Chen and Jiaxing Xu and Jiazheng Xu and Jing Chen and Jinghao Lin and Jinhao Chen and Jinjiang Wang and Junjie Chen and Leqi Lei and Letian Gong and Leyi Pan and Mingdao Liu and Mingde Xu and Mingzhi Zhang and Qinkai Zheng and Ruiliang Lyu and Shangqin Tu and Sheng Yang and Shengbiao Meng and Shi Zhong and Shiyu Huang and Shuyuan Zhao and Siyan Xue and Tianshu Zhang and Tianwei Luo and Tianxiang Hao and Tianyu Tong and Wei Jia and Wenkai Li and Xiao Liu and Xiaohan Zhang and Xin Lyu and Xinyu Zhang and Xinyue Fan and Xuancheng Huang and Yadong Xue and Yanfeng Wang and Yanling Wang and Yanzi Wang and Yifan An and Yifan Du and Yiheng Huang and Yilin Niu and Yiming Shi and Yu Wang and Yuan Wang and Yuanchang Yue and Yuchen Li and Yusen Liu and Yutao Zhang and Yuting Wang and Yuxuan Zhang and Zhao Xue and Zhengxiao Du and Zhenyu Hou and Zihan Wang and Peng Zhang and Debing Liu and Bin Xu and Juanzi Li and Minlie Huang and Yuxiao Dong and Jie Tang},
      year={2026},
      eprint={2507.01006},
      archivePrefix={arXiv},
      primaryClass={cs.CV},
      url={https://arxiv.org/abs/2507.01006}, 
}

@misc{huang2026step3vl10btechnicalreport,
      title={STEP3-VL-10B Technical Report}, 
      author={Ailin Huang and Chengyuan Yao and Chunrui Han and Fanqi Wan and Hangyu Guo and Haoran Lv and Hongyu Zhou and Jia Wang and Jian Zhou and Jianjian Sun and Jingcheng Hu and Kangheng Lin and Liang Zhao and Mitt Huang and Song Yuan and Wenwen Qu and Xiangfeng Wang and Yanlin Lai and Yingxiu Zhao and Yinmin Zhang and Yukang Shi and Yuyang Chen and Zejia Weng and Ziyang Meng and Ang Li and Aobo Kong and Bo Dong and Changyi Wan and David Wang and Di Qi and Dingming Li and En Yu and Guopeng Li and Haiquan Yin and Han Zhou and Hanshan Zhang and Haolong Yan and Hebin Zhou and Hongbo Peng and Jiaran Zhang and Jiashu Lv and Jiayi Fu and Jie Cheng and Jie Zhou and Jisheng Yin and Jingjing Xie and Jingwei Wu and Jun Zhang and Junfeng Liu and Kaijun Tan and Kaiwen Yan and Liangyu Chen and Lina Chen and Mingliang Li and Qian Zhao and Quan Sun and Shaoliang Pang and Shengjie Fan and Shijie Shang and Siyuan Zhang and Tianhao You and Wei Ji and Wuxun Xie and Xiaobo Yang and Xiaojie Hou and Xiaoran Jiao and Xiaoxiao Ren and Xiangwen Kong and Xin Huang and Xin Wu and Xing Chen and Xinran Wang and Xuelin Zhang and Yana Wei and Yang Li and Yanming Xu and Yeqing Shen and Yuang Peng and Yue Peng and Yu Zhou and Yusheng Li and Yuxiang Yang and Yuyang Zhang and Zhe Xie and Zhewei Huang and Zhenyi Lu and Zhimin Fan and Zihui Cheng and Daxin Jiang and Qi Han and Xiangyu Zhang and Yibo Zhu and Zheng Ge},
      year={2026},
      eprint={2601.09668},
      archivePrefix={arXiv},
      primaryClass={cs.CV},
      url={https://arxiv.org/abs/2601.09668}, 
}

@misc{an2026llavaonevision2nextgenerationperceptualintelligence,
      title={LLaVA-OneVision-2: Towards Next-Generation Perceptual Intelligence}, 
      author={Xiang An and Yin Xie and Feilong Tang and Yunyao Yan and Huajie Tan and Didi Zhu and Changrui Chen and Xiuwei Zhao and Bin Qin and Kaicheng Yang and Yifei Shen and Yuanhan Zhang and Kaichen Zhang and Wenkang Zhang and Zheng Cheng and Nansen Zhang and Chunsheng Wu and Chunjiang Ge and Zimin Ran and Dehua Song and Chunyuan Li and Shikun Feng and Ming Hu and Zhangquan Chen and Junbo Niu and Bo Li and Ziyong Feng and Ziwei Liu and Zongyuan Ge and Jiankang Deng},
      year={2026},
      eprint={2605.25979},
      archivePrefix={arXiv},
      primaryClass={cs.CV},
      url={https://arxiv.org/abs/2605.25979}, 
}

@misc{Gemma-4-E4B-it,
      title={Gemma 4 Technical Report}, 
      author={Gemma Team and Sherif El Abd and Vaibhav Aggarwal and Robin Algayres and Alek Andreev and Olivier Bachem and Ian Ballantyne and Cormac Brick and Victor Cărbune and Michelle Casbon and Mayank Chaturvedi and Aditya Chawla and Victor Cotruta and Alice Coucke and Phil Culliton and Robert Dadashi and Lucas Dixon and Mohamed Elhawaty and Utku Evci and Clément Farabet and Johan Ferret and Filippo Galgani and Sertan Girgin and Jean-Bastien Grill and Maarten Grootendorst and Jiaxian Guo and Cassidy Hardin and Yanzhang He and Steven M. Hernandez and Omri Homburger and Léonard Hussenot and Juyeong Ji and Armand Joulin and Aishwarya Kamath and Parnian Kassraie and Olivier Lacombe and Preethi Lahoti and Gaël Liu and Gus Martins and Luciano Martins and Tatiana Matejovicova and Ramona Merhej and Nikola Momchev and Sneha Mondal and Ryan Mullins and Sindhu Raghuram Panyam and Shreya Pathak and Sarah Perrin and André Susano Pinto and Etienne Pot and Angéline Pouget and Alexandre Ramé and Sabela Ramos and Douglas Reid and David Rim and Morgane Rivière and Karsten Roth and Louis Rouillard and Omar Sanseviero and Pier Giuseppe Sessa and Shane Settle and Danila Sinopalnikov and Sara Smoot and Piotr Stanczyk and Andreas Steiner and Lawrence Stewart and Ilya Tolstikhin and Michael Tschannen and Anton Tsitsulin and Nino Vieillard and Renjie Wu and Pingmei Xu and Haichuan Yang and Edouard Yvinec and Biao Zhang and Li Zhang and Joe Zou and Nicolas Aagnes and Abdelrahman Abdelhamed and Jakub Adamek and Shivani Agrawal and Shubham Agrawal and Ibrahim Alabdulmohsin and Jean Baptiste Alayrac and Uri Alon and Chandramouli Amarnath and Ankesh Anand and Chrysovalantis Anastasiou and Setareh Ariafar and François-Xavier Aubet and Kyriakos Axiotis and Federico Barbero and Joelle Barral and Alexei Bendebury and Urs Bergmann and Stanley Bileschi and Kat Black and Mathieu Blondel and Sebastian Borgeaud and Arthur Bražinskas and Ryan Burnell and Robert Busa-Fekete and Mu Cai and Daniele Calandriello and Glenn Cameron and Charlotte Caucheteux and Rahma Chaabouni and Garima Chadha and Jetha Chan and Blake Jianhang Chen and Jesse Chen and Lin Chen and Xu Chen and Derek Cheng and Tzu-hsiang Chien and Nikolai Chinaev and Yi Chou and Zhaohui Chu and Benjamin Coleman and Pooja Consul and Sam Conway-Rahman and Scott Crowell and Dylan Cutler and Vivek Dani and Samira Daruki and Anil Das and Daniel Deutsch and Nishanth Dikkala and Li Ding and Qiuhan Ding and Shenil Dodhia and Konstantin Donhauser and Tulsee Doshi and Anca Dragan and Alex Druinsky and Sahil Dua and Zoltan Egyed and Danielle Eisenbud and Daniel Eppens and Cindy Fan and Bahare Fatemi and Yassir Fathullah and Vlad Feinberg and Milen Ferev and Sebastian Flennerhag and Takumi Fujimoto and João Gabriel Oliveira and Isaac Galatzer-Levy and João Gante and Simon Geisler and Soham Ghosal and Antonious M. Girgis and Tamara von Glehn and Alec Go and Alhaad Gokhale and Alex Grills and Yiming Gu and Mayank Gupta and Pramod Gupta and Guru Guruganesh and Raia Hadsell and Hamza Harkous and Jitendra Harlalka and Demis Hassabis and Anja Hauth and Joe Heyward and Arian Hosseini and Chih-Yang Hsia and I-Hung Hsu and Xiaopeng Huang and Yangsibo Huang and Kevin Hui and Adrian Hutter and Te I and Fotis Iliopoulos and Advait Jain and Ganesh Jawahar and Ziwei Ji and Qilin Jin and Melvin Johnson and Kandarp Joshi and Arun Kandoor and Wang-Cheng Kang and Koray Kavukcuoglu and Mehran Kazemi and Kathleen Kenealy and Amr Khalifa and Phoebe Kirk and Ivan Korotkov and Suraj Kothawade and Vitaly Kovalev and Neel Kovelamudi and Adam Kraft and Ravin Kumar and Vivek Kumar and Harish Kuppam and Justin Lannin and Chen-Yu Lee and Seungji Lee and Dmitry Lepikhin and Alon Levkovitch and Dongdong Li and Qiujia Li and Valentin Liévin and Ethan Lin and Ziqian Lin and Casper Liu and Tianlin Liu and Tianqi Liu and Xin Liu and Ivan Lobov and Mayank Lunayach and Min Ma and Gagan Madan and Andrii Maksai and Eric Malmi and Michal Matuszak and Daniel McDuff and Gaurav Menghani and Maciej Mikuła and Daniil Mirylenka and Karolis Misiunas and Vedant Misra and Andreea Mitran and Kareem Mohamed and Maksim Mukha and Eric Noland and James O'Donnell and Brendan O'Donoghue and Kate Olszewska and Bernett Orlando and Wanqiong Pan and Rina Panigrahy and Unnati Parekh and Nicolas Perez-Nieves and Chunjong Park and Eric Paskie and Liqian Peng and Bryce Petrini and Slav Petrov and Jonas Pfeiffer and Bilal Piot and Martyna Plomecka and Siim Poder and Octavio Ponce and Arijit Pramanik and David Racz and Anish Rajan and Michelle Ramanovich and Anand Rao and Marvin Ritter and Vitor Rodrigues and Evan Rosen and Mikołaj Rybiński and Noveen Sachdeva and Michaël E. Sander and Rohit Sathyanarayana and Sagar Savla and Samuel Schmidgall and Tal Schuster and George Scrivener and Benoit Seguin and Andrew Sellergren and Aliaksei Severyn and Izhak Shafran and Dhruv Shah and Bobak Shahriari and Yuan Shangguan and Ashish Shenoy and Pradeep Shenoy and Rakesh Shivanna and Pauline Sho and Lucas Spangher and Wojciech Stokowiec and Tim Strother and Yao Su and Yinghao Sun and Mukund Sundararajan and Andrea Tacchetti and Mor Hazan Taege and Pouya Tafti and Jean Tarbouriech and Chetan Tekur and Shantanu Thakoor and Rahul Thapa and Madeleine Traverse and Lenart Treven and Tao Tu and Chien Te Tung and Çağlar Ünlü and Petar Veličković and Malini Pooni Venkat and Sagar Gubbi Venkatesh and Vidya Venkiteswaran and Francesco Visin and Alex Vitvitskyi and Kiran Vodrahalli and Weiyi Wang and Xin Wang and Tris Warkentin and Jan Wassenberg and John Wieting and Cindy Wu and Lechao Xiao and Hao Xu and Yuhui Xu and Fuzhao Xue and Arun Yadav and Jun Yan and Antoine Yang and Lin Yang and Ming-Hsuan Yang and Ziyu Ying and Jae Hyeon Yoo and Morteza Zadimoghaddam and Sajjad Zafar and Fred Zhang and Jiageng Zhang and Jianyi Zhang and Xiaofan Zhang and Chao Zhao and David Zhou and Chen Zou},
      year={2026},
      eprint={2607.02770},
      archivePrefix={arXiv},
      primaryClass={cs.CL},
      url={https://arxiv.org/abs/2607.02770}, 
}

@misc{clark2026molmo2openweightsdata,
      title={Molmo2: Open Weights and Data for Vision-Language Models with Video Understanding and Grounding}, 
      author={Christopher Clark and Jieyu Zhang and Zixian Ma and Jae Sung Park and Mohammadreza Salehi and Rohun Tripathi and Sangho Lee and Zhongzheng Ren and Chris Dongjoo Kim and Yinuo Yang and Vincent Shao and Yue Yang and Weikai Huang and Ziqi Gao and Taira Anderson and Jianrui Zhang and Jitesh Jain and George Stoica and Winson Han and Ali Farhadi and Ranjay Krishna},
      year={2026},
      eprint={2601.10611},
      archivePrefix={arXiv},
      primaryClass={cs.CV},
      url={https://arxiv.org/abs/2601.10611}, 
}

@misc{qwen3.6-27b,
    title  = {{Qwen3.6-27B}: Flagship-Level Coding in a {27B} Dense Model},
    author = {{Qwen Team}},
    month  = {April},
    year   = {2026},
    url    = {https://qwen.ai/blog?id=qwen3.6-27b}
}

@misc{mistral_small_3_1_24b_instruct,
    author = {{Mistral AI}},
    title = {{Mistral Small 3.1}},
    year = {2025},
    month = {March},
    url = {https://mistral.ai/news/mistral-small-3-1/},
    urldate = {2025-12-05},
  }

@misc{llama32_11b_vision_instruct,
    author = {{Meta AI}},
    title = {{Llama 3.2: Revolutionizing edge AI and vision with open, customizable
    models}},
    year = {2024},
    month = {September},
    url =
    {https://ai.meta.com/blog/llama-3-2-connect-2024-vision-edge-mobile-devices/},
    urldate = {2025-12-05},
  }
\
\clearpage
\appendix


\section{Ethical Statement}
All data used in this study were obtained either from publicly available sources or through experiments conducted by the authors. The SAFIRE benchmark was curated to advance research on safety-critical visual reasoning in fire and smoke scenarios and will be released for non-commercial research use. All pretrained models (e.g., Qwen-VL, InternVL) were employed in accordance with their respective open-source licenses. We used AI tool (e.g. Codex (GPT-5.4) ) for data cleaning, and improving the language clarity of the main content; all conceptual and scientific content was solely verified by the authors.

\section{Data Collection and Cleaning}\label{app:dataCleaning}
\subsection{Dataset Collection}
Images representing diverse fire and smoke scenarios were collected using a keyword-based search strategy across multiple web sources. To ensure high relevance and quality, we employed a multi-factor filtering pipeline throughout the collection process.

The first filtering step involved a strict keyword conjunction mechanism. For instance, to retrieve images for the ``Open grassland fire" scenario, we used the combined query terms ``grassland", ``farm", ``flatland" and ``burning". Only images associated with all specified keywords were retained, thereby reducing semantic noise and increasing specificity. Additionally, a manually defined threshold parameter $\sigma$ was introduced as a stopping criterion for automated webpage crawling. For example, in a web query structured as `` \textit{https://webname/search/term1\&term2\&term3/page =[1:$\sigma$]} ", the crawler halted once the results began returning irrelevant content, preventing excessive inclusion of low-quality or unrelated samples.

Together, these two automated filters eliminated over 37\% of irrelevant images prior to manual review, significantly improving the dataset's precision and reducing the overall cleaning burden in downstream stages.

\subsection{Source Provenance}
Table~\ref{tab:source_provenance} summarizes the provenance of the raw candidate data before filtering. We combine published research datasets, public dataset hubs, public web platforms, authorized field data, and controlled combustion experiments to improve scenario diversity, with video-based sources being sampled into candidate frames before duplicate removal and semantic filtering. The controlled experiment set up was built in Sas Al Nakhl Campus of Khalifa University during Nov. 2023 to Mar. 2024 and the entire data collection protocol was approved by the university's Ethics Review Board. We reiterate that the resulting benchmark is for non-commercial research purposes only.

\begin{table*}[t]
\centering
\footnotesize
\caption{\textbf{Source provenance of raw candidate data before filtering.} Counts are measured before the multi-factor filtering and manual inspection stages.}
\label{tab:source_provenance}
\setlength{\tabcolsep}{4pt}
\renewcommand{\arraystretch}{1.24}
\setlength{\extrarowheight}{1.5pt}
\begin{tabular}{@{}>{\raggedright\arraybackslash}p{0.16\textwidth} >{\raggedright\arraybackslash}p{0.30\textwidth} >{\raggedleft\arraybackslash}p{0.13\textwidth} >{\raggedright\arraybackslash}p{0.35\textwidth}@{}}
\toprule
\textbf{Source group} & \textbf{Source(s)} & \textbf{Raw data} & \textbf{Access} \\
\midrule
Published datasets & \textit{BowFire}~\cite{chino2015bowfire} (466); \textit{UniInd-FireSmoke}~\cite{UniInd} (3{,}789) & 4{,}255 samples & Public research datasets. We cite the original sources and retain their research-use license constraints. \\
\addlinespace[3pt]
Roboflow datasets & 15 public datasets, each containing 1{,}270--2{,}657 samples & 25{,}879 samples & Public dataset sources. \\
\addlinespace[4pt]
Public platforms & YouTube (9{,}743); Shutterstock (9{,}356); Getty (9{,}589); TikTok (15{,}712); Bilibili (19{,}145); iStock (26{,}869) & 90{,}414 images & Images or frames collected and filtered for non-commercial research use. SAFIRE is released for research purposes only. \\
\addlinespace[4pt]
Field data & Gas-flare samples & 2{,}271 samples & Collected with authorized field access. \\
\addlinespace[3pt]
Controlled experiments & 80 combustion experiments & About 12{,}600 frames & Collected by varying O$_2$, CH$_4$, C$_2$H$_6$ composition and flow velocity to generate diverse fire and smoke behaviors; \\
\bottomrule
\end{tabular}
\end{table*}

The raw provenance counts are approximate candidate records and may include similar images or low-relevance examples. After source-level filtering, semantic verification, and manual inspection, these candidates are reduced to the final 83K high-quality SAFIRE images described in Table~\ref{tab:image_qna_counts}.

\subsection{Dataset Cleaning}
Following data collection, we conducted a manual curation phase to refine the scenario selection. Each candidate scenario was assessed based on two main criteria: (1) overall image quality and noise level, and (2) semantic relevance to fire and smoke analysis. Several scenarios were discarded due to high false-positive rates or conceptual redundancy.

For example, the ``Satellite Massive Fire'' scenario was excluded due to frequent misclassifications involving laminar flow, haze, or cloud cover. Similarly, the following scenarios were removed for specific reasons:

\begin{itemize}
    \item \textbf{Lightning Strike}: Predominantly returned synthetic computer-generated imagery (CGI) with low diversity and high semantic ambiguity (e.g., “lightning strike” may refer to military operations).
    
    \item \textbf{Welding Operation}: Often featured icons or close-up shots of machinery or workers, lacking sufficient visual focus on sparks or flames.
    
    \item \textbf{Electrical Fire}: Substantially overlapped with the \emph{Home Fire} category, as most images depicted socket-related incidents already represented in the residential context.
    
    \item \textbf{Traditional Festival Rituals}: Visually resembled \emph{Campfire} or \emph{Bonfire} scenes, adding little scenario diversity.
\end{itemize}

After finalizing the 20 scenario categories and the initial data collection, we applied an \textbf{automation-manual combined cleaning process} across all retained images to remove remaining irrelevant or low-quality samples. This reduced the dataset from an initial 135.4K auto-collected images to 91K (after GPT-5.4 semantic match examination), and finally \textbf{83K high-quality images} (after manual inspection).


As a result, only \textbf{61.2\%} of the initial data was retained in the final SAFIRE benchmark. This rigorous filtering yields a cleaner, more diverse, and semantically coherent dataset, providing a solid foundation for robust vision-language research in fire and smoke understanding. SAFIRE is released for research use only, and a summary of final statistics and dataset categorization is provided in Table~\ref{tab:image_qna_counts}.

\section{Dataset Composition \& Structuring} \label{app:dataComposition}

\definecolor{myBG}   {RGB}{248,248,248}
\definecolor{myGreen}{RGB}{230,255,230}
\definecolor{myCyan} {RGB}{220,245,255}
\definecolor{myYellow}{RGB}{255,245,180}


\subsection{Annotation Structure and Benchmark Objectives}
SAFIRE comprises two complementary annotation branches, each tailored to evaluate distinct multimodal reasoning capabilities:

\begin{itemize}
    \item \textbf{Multi-Choice Visual Question Answering (MCVQA):} This branch is designed to assess the fine-grained reasoning abilities of MLLMs. It includes 193K questions spanning 10 reasoning dimensions (e.g., context, causality, behavior), targeting high-level understanding of fire and smoke scenes beyond simple recognition.

    \item \textbf{Image-Caption (IC) Pairs:} This branch supports the evaluation of multimodal encoders, particularly in zero-shot and retrieval settings. Each of the 83K images is paired with a carefully constructed caption capturing key semantic attributes, including actors, objects, fire-smoke states, and safety cues.
\end{itemize}

These branches enable comprehensive benchmarking across reasoning depth and modality alignment, supporting both autoregressive MLLMs and encoder-based architectures. Each image is also annotated with a fire category and scene label to facilitate fine-grained scenario analysis.

\subsection{Prompting Strategy and Instruction Design} 

All annotations, including both MCVQA and IC pairs, are generated simultaneously using a structured prompting pipeline built on the Qwen2.5-VL-72B-Instruct \cite{bai2025qwen25vltechnicalreport} vision language model. A standardized instruction defines the expected output format and covers 10 reasoning dimensions such as scenario type, spatial localization, fire-smoke interaction, and human activity.


To minimize hallucinations, the model is explicitly instructed to \textit{“strictly follow the image content when generating questions and answers”}. In ambiguous situations, a “None of the above” option is included, allowing the model to abstain when none of the listed answers is correct. The following boxes illustrate the default system instruction and captioning guidelines used during annotation:

\vspace{1mm}
\begin{tcolorbox}[colback=myBG, colframe=black,
                  arc=2mm, boxrule=0.5pt,
                  left=2mm,right=2mm,top=1mm,bottom=1mm,
                  breakable, title=\textbf{Default Instruction}]
``You are an advanced VLM tasked with producing detailed captions for fire‑related images, each from one of 20 predefined fire-related scenarios..." \\
...... \\
``... Generate captions that thoroughly describe visible fire and smoke elements. Include sufficient details for the 10 evaluation dimensions... Form ground-truth QA pairs to assess reasoning capabilities."
\end{tcolorbox}

\begin{tcolorbox}[colback=myBG, colframe=black,
                  arc=2mm, boxrule=0.5pt,
                  left=2mm,right=2mm,top=1mm,bottom=1mm,
                  breakable, title=\textbf{Captioning Guidelines}]
1. Context Awareness: Confirm the fire-related context and avoid symbolic or irrelevant images. \\
2. Visual Description: Describe key attributes of fire and smoke, surrounding context, and activity. \\
3. Output Format: Produce a structured JSON object with captions, key indicators, and MCVQA entries.
\end{tcolorbox}

\subsection{MCVQA Design and Reasoning Dimensions}

Each selected MCVQA-active image is paired with approximately 20 multiple-choice questions targeting multi-level visual and contextual reasoning. These are categorized as follows:

\begin{tcolorbox}[colback=myBG, colframe=black,
                  arc=2mm, boxrule=0.5pt,
                  left=2mm,right=2mm,top=1mm,bottom=1mm,
                  breakable, title=\textbf{QA Instructions (Q1-Q20)}]

\textbf{Scene-Level Evaluation (Q1)}: Scenario classification, e.g., “What kind of fire/smoke is this?” \\
\textbf{Spatial \& Instance-Level (Q2-Q4)}: Fire position (left/center/right), number of fire sources, danger level. \\
\textbf{Recognition (Q5-Q10)}: Smoke presence, volume, color, flame visibility, time of day, negative cases. \\
\textbf{Reasoning \& Context (Q11-Q20)}: Fire source, human presence, firefighting activity, trend prediction, emergency response, and semantic ambiguity.
\end{tcolorbox}

\subsection{Structured Output Format}

All annotations conform to a unified JSON schema, consisting of the image ID, caption, key indicators, and MCVQA entries. This format ensures compatibility with downstream tools and promotes reproducibility in multimodal benchmarks.

\begin{tcolorbox}[colback=myBG, colframe=black,
                  arc=2mm, boxrule=0.5pt,
                  left=2mm,right=2mm,top=1mm,bottom=1mm,
                  breakable, title=\textbf{Output JSON Skeleton}]
\small
\begin{verbatim}
{ 
  "Image_name": "...", "caption": "...",
  "Key Indicators": {
    "pos": "...", "danger level": "...", 
    "Fire trend": "..."
  },
  "QA_pairs": [
    {"Question 1": "...", "Options": [...], 
    "Answer": "..."},
    {"Question 2": "...", "Options": [...], 
    "Answer": "..."},
    ...
  ]
}
\end{verbatim}
\end{tcolorbox}

\subsection{Handling MCVQA Failures}\label{sec:MCVQA}
During caption generation, a subset of samples failed to yield valid JSON. These failures were categorized as: (1) \textbf{complete failures}, with no output returned, and (2) \textbf{partial failures}, where captions were generated but no MCVQA pairs. In partial cases, captions were retained and included in the CLIP branch, enriching the pool of aligned image-text pairs.


\begin{figure}[t]
\centering
\resizebox{\columnwidth}{!}{  
\begin{tikzpicture}[node distance=1.5cm and 2.2cm]

\tikzstyle{startstop} = [rectangle, rounded corners, minimum width=3cm, minimum height=1cm,text centered, draw=black, fill=blue!10]
\tikzstyle{process} = [rectangle, minimum width=3cm, minimum height=1cm, text centered, draw=black, fill=gray!10]
\tikzstyle{decision} = [diamond, aspect=2, text centered, draw=black, fill=orange!10, inner sep=1pt]
\tikzstyle{arrow} = [thick,->,>=stealth]

\node (start) [startstop] {Image \& QA Pairs in .parquet Format};
\node (load) [process, below of=start] {Load Image \& Metadata};
\node (build) [process, below of=load] {Build Prompt with Question-Options};
\node (encode)  [process, below of=build]  {Encode Image \& Text};
\node (fusion)  [process, below of=encode] {Cross-Modal Fusion};
\node (generate)[process, below of=fusion] {Question with Unique ID $\leftarrow$ MLLMs' Responses};

\node (extract) [process, below of=generate] {Extract Letters (A--D) across MLLMs using ``Regex"};
\node (compare) [process, below of=extract] {Compare with Ground Truth};

\node (check) [decision, below of=compare, yshift=-0.2cm] 
  {$\geq 2$ MLLMs $\neq$ Ground Truth?};

\node (update) [process, below left=1.7cm and 2.2cm of check] {Manual Check the Question};
\node (skip) [process, below right=1.7cm and 2.2cm of check] {Skip Sample};

\node (end) [startstop, below=3cm of check] {Cleaned MCVQA after Major Voting};

\draw [arrow] (start) -- (load);
\draw [arrow] (load) -- (build);
\draw [arrow] (build) -- (encode);
\draw [arrow] (encode) -- (fusion);
\draw [arrow] (fusion) -- (generate);

\draw [arrow] (generate) -- (extract);

\draw [arrow] (extract) -- (compare);
\draw [arrow] (compare) -- (check);
\draw [arrow] (check) -- node[above left] {\small Yes} (update);
\draw [arrow] (check) -- node[above right] {\small No} (skip);
\draw [arrow] (update) |- (end);
\draw [arrow] (skip) |- (end);

\end{tikzpicture}
} 
\caption{\textbf{Majority voting pipeline for cleaning multi-scenario MCVQA answers from multiple MLLMs.}}
\label{fig:vqa-clean-pipe}
\end{figure}
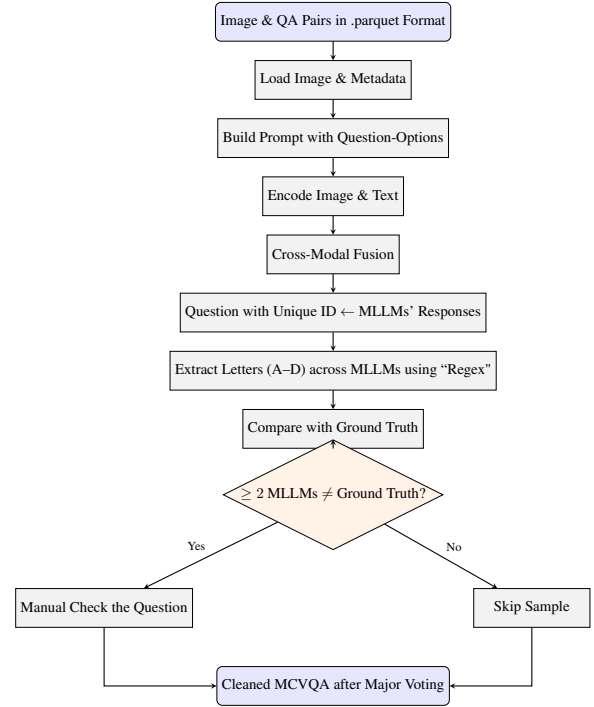

\begin{table*}[t]
\centering
\small
\caption{\textbf{Residual annotation error analysis.} Error categories are computed over the 12.9\% incorrect cases identified in the 2,800-sample human audit.}
\label{tab:annotation_error_types}
\setlength{\tabcolsep}{4pt}
\renewcommand{\arraystretch}{1.18}
\begin{tabular}{p{0.29\textwidth} r p{0.56\textwidth}}
\toprule
\textbf{Error type} & \textbf{Percent (\%)} & \textbf{Example} \\
\midrule
Fire/smoke visibility and visual attributes & 24.5 & \texttt{campf\_654.jpg}: a pile of unlit wood was labeled as visible flame. \\
Semantic mismatch & 20.8 & \texttt{volca\_308.jpg}: a white steam plume was labeled as molten lava glow. \\
Answer format error & 17.2 & \texttt{rocke\_424.jpg}: the ground-truth text was ``Emergency response'', but it used the wrong option label. \\
Context inference error & 14.2 & \texttt{fores\_557.jpg}: a firefighter wearing a black uniform in a burned forest was labeled as no people and no active fire, which is only partially correct. \\
Localization/counting error & 9.9 & \texttt{firew\_213.jpg}: a cartoon of fireworks was labeled as having more than two real fire sources. \\
Real vs. symbolic error & 8.4 & \texttt{home\_419.jpg}: a fire icon was labeled as literal fire. \\
Spatial relation error & 5.1 & \texttt{rocke\_93.jpg}: an aircraft smoke trail was labeled as smoke rising above fire, while it actually trails behind the aircraft. \\
\bottomrule
\end{tabular}
\end{table*}

\subsection{GPT-5.4 Verification and Majority Voting for MCVQA Correction}\label{app:quality control}
\begin{table}[h]
\centering
\small
\caption{\textbf{Inter-model consistency in majority voting.}}
\label{tab:majority_vote_consistency}
\setlength{\tabcolsep}{6pt}
\renewcommand{\arraystretch}{1.12}
\begin{tabular}{l r r}
\toprule
\textbf{Votes} & \textbf{\# Samples} & \textbf{Rate} \\
\midrule
3/3 same answer & 101{,}332 & 52.53\% \\
2/3 same answer & 85{,}958 & 44.56\% \\
3 different answers & 5{,}614 & 2.91\% \\
\bottomrule
\end{tabular}
\end{table}

\begin{table}[h]
\centering
\small
\caption{\textbf{Model-wise agreement with the original answer.} No single verifier dominates the voting process.}
\label{tab:verifier_agreement}
\setlength{\tabcolsep}{8pt}
\renewcommand{\arraystretch}{1.12}
\begin{tabular}{l c}
\toprule
\textbf{Verifier model} & \textbf{Agreement} \\
\midrule
 Mistral-Small-3.1-24B-Instruct & 56.67\% \\
  Gemma-3-12B-it & 55.46\% \\
  Llama-3.2-11B-Vision-Instruct & 52.37\% \\
\bottomrule
\end{tabular}
\end{table}
To ensure the quality of MCVQAs, we adopt a GPT-5.4-assisted cleaning pipeline combined with majority voting across multiple MLLMs. 

The verification process consists of two rounds of refinement. In the first round, we design scenario-specific prompts to make the model aware of potential contextual biases and critical visual cues for each question. For example, in the \textit{Firework} scenario, the model is explicitly instructed to recognize that fire may dominate the scene, carefully count distinct bursts, and avoid over-counting spark fragments. 

In the second round, GPT-5.4 performs logical consistency checking to identify contradictions between image content and generated answers. For instance, if an image contains only fire without visible smoke, questions regarding the ``spatial relationship between fire and smoke" should be answered as \textit{N/A}, and if the visible evidence indicates a fire with only fireplace/wood/bricks, then a “high danger” label is usually inconsistent and should be corrected. 

After GPT-5.4 refinement, we further apply majority voting as a disagreement detector. As illustrated in Fig.~\ref{fig:vqa-clean-pipe}, instead of using majority voting as an automatic relabeling mechanism, we use this method to signal potential incorrect data where most models fail. We use three non-Qwen verifier models that are independent of the Qwen annotation generator: LLaMA-3.2-11B, Mistral-3.1-24B, and Gemma-3-12B. For each question, model predictions (options A--D) are extracted using regular expressions and compared with the original annotated answer. Samples are flagged for manual review when the three verifiers produce no clear majority, or when the majority vote contradicts the annotated answer; otherwise, they are automatically accepted. This strategy efficiently prioritizes human inspection toward potentially flawed annotations while avoiding direct dependence on a single verifier.

\begin{table*}[t]
\centering
\small
  \caption{\textbf{Full-benchmark and manually audited subset accuracy.} Results compare model accuracy on the complete
  SAFIRE MCVQA set and the manually corrected 2,800-MCVQA set.}
  
\label{tab:clean_subset_effect}
\setlength{\tabcolsep}{5pt}
\renewcommand{\arraystretch}{1.15}
\begin{tabular}{l c c c}
\toprule
\textbf{Model} & \textbf{Original Full-Set Acc.} & \textbf{Audited-Subset Acc.} & \textbf{Difference} \\
\midrule
Qwen3.5-9B non-thinking & 64.89\% & 68.98\% & +4.09 \\
Step3-VL-10B & 63.39\% & 67.96\% & +4.57 \\
Qwen3.5-9B thinking & 62.17\% & 67.41\% & +5.24 \\
InternVL3.5-8B & 58.37\% & 63.73\% & +5.36 \\
LLaVA-OneVision-2-8B & 61.61\% & 64.30\% & +2.69 \\
GLM-4.1V-9B-Base & 59.70\% & 62.90\% & +3.20 \\
Gemma-4-E4B-it & 59.30\% & 60.60\% & +1.30 \\
Molmo2-8B & 56.92\% & 60.29\% & +3.37 \\
\bottomrule
\end{tabular}
\end{table*}

As shown in Table~\ref{tab:majority_vote_consistency}, the three verifier models produce a clear majority in 97.09\% of MCVQA samples. Only 2.91\% of samples produce three different answers, indicating that complete inter-model disagreement is rare. The final manually double-reviewed set from this process consists of three mutually exclusive subsets: (1) cases where the three verifier models give three different answers (5,614 samples), (2) cases where the original annotated answer differs from the 2/3 majority vote (1,308 samples), and (3) cases where the original annotated answer differs from the 3/3 unanimous vote (215 samples). In total, 7,137 MCVQA samples, corresponding to 3.70\% of the full MCVQA set, were double-reviewed through the majority-voting process.

Table~\ref{tab:verifier_agreement} further reports model-wise agreement with the original annotated answers. The three agreement rates suggest that the voting procedure does not over-rely on a single dominant verifier model. Moreover, the pipeline can identify ambiguous questions where model responses fall outside the predefined options, indicating unclear or inconsistent labels that require further investigation.

\section{Annotation Error Analysis}
\label{app:annotation_error_analysis}

Building on the human verification described in Appendix~\ref{app:quality control}, we further analyze the incorrect portion of the 2,800 manually audited MCVQA samples. Table~\ref{tab:annotation_error_types} summarizes the main residual error types from the remaining 12.9\% annotation error rate. Most errors are correctable visual-grounding ambiguities, semantic mismatches, or answer-format inconsistencies.


 \paragraph{Robustness under manually audited labels.}

   To assess whether residual annotation noise could affect the benchmark conclusions, we further evaluated an extended
  family of models on the manually audited 2,800-MCVQA subset. This subset was manually corrected after the annotation audit and provides a higher-confidence reference set for examining whether the observed model trends remain stable under cleaner labels. We use this analysis as a robustness check rather than a precise estimation of the isolated effect of label correction because the audited subset is not identical to the full benchmark. As shown in Table~\ref{tab:clean_subset_effect}, models obtain accuracies that are 1.30 to 5.36 percentage points higher on the audited subset than on the full SAFIRE MCVQA benchmark. More importantly, the relative ordering of models remains broadly consistent, and all models still perform well below human reference performance. These results suggest that residual annotation noise may affect absolute scores, but it does not overturn the main conclusion that current MLLMs still
  struggle with SAFIRE's fire-smoke reasoning tasks.

\section{Dimension Correlation Analysis}
\label{app:dimension_correlation}

To further analyze the structure of SAFIRE, we measure the association between different evaluation dimensions. We first compute correlations at the question-template level. Each answer is converted to its multiple-choice option label (A--D), and Cram\'er's~V is computed for every pair of question templates. We then aggregate these scores to the dimension level: for any two dimensions, their correlation is the average Cram\'er's~V across all question-template pairs drawn from the two dimensions. Values near 0 indicate weak association, while values near 1 indicate strong overlap. 

Table~\ref{tab:dimension_correlation_extremes} selectively reports the strongest and weakest cross-dimension associations, where we can find even the strongest association is moderate: \textit{Classification} and \textit{People Emotional Response} reach 0.410, which is expected because the type of fire or smoke scene often influences how people are likely to react. The second strongest association, between \textit{Target Counting} and \textit{Position Identification} (0.407), is also intuitive because both rely on locating salient fire or smoke sources in the image. Similarly, \textit{Fire/Smoke Intention} and \textit{People Emotional Response} are associated (0.365), since controlled, accidental, and natural fire contexts imply different human reactions.

\begin{table}[t]
\centering
\small
\caption{\textbf{Strongest and weakest cross-dimension associations.} Abbreviations: Class. = Classification, Emot. Resp. = People Emotional Response, Pos. Ident. = Position Identification, Intention = Fire/Smoke Intention, Human Pres. = Human Presence, Spatial Corr. = Fire--smoke Spatial Correlation.}
\label{tab:dimension_correlation_extremes}
\setlength{\tabcolsep}{3pt}
\renewcommand{\arraystretch}{1.12}
\begin{tabular}{@{}l p{0.24\linewidth} p{0.29\linewidth} c@{}}
\toprule
\textbf{Type} & \textbf{Dim. A} & \textbf{Dim. B} & \textbf{Mean V} \\
\midrule
High-1 & Class. & Emot. Resp. & 0.410 \\
High-2 & Counting & Pos. Ident. & 0.407 \\
High-3 & Intention & Emot. Resp. & 0.365 \\
\midrule
Low-1 & Human Pres. & Pos. Ident. & 0.030 \\
Low-2 & Human Pres. & Spatial Corr. & 0.040 \\
Low-3 & Counting & Human Pres. & 0.060 \\
\bottomrule
\end{tabular}
\end{table}


The weakest correlations are substantially lower. \textit{Human Presence} has very weak association with \textit{Position Identification} (0.030), \textit{Fire--smoke Spatial Correlation} (0.040), and \textit{Target Counting} (0.060), indicating that whether people appear in an image is largely independent of where fire or smoke is located and how many sources are visible. The full correlation analysis also shows low-to-moderate associations for other semantically distinct pairs, such as \textit{Fire--smoke Spatial Correlation} and \textit{General Reasoning} (0.190). Overall, these results indicate that SAFIRE dimensions are semantically related where expected, but remain sufficiently distinct to support multi-dimensional evaluation.


\section{Scenario-Prior Ablation Study}
\label{app:scenario_prior_ablation}
In the MCVQA evaluation, certain scenarios may be correlated with particular answers. For example, stove scenes may be associated with controlled cooking, whereas residential-fire scenes may be associated with danger or emergencies. Such scenario–answer correlations could provide models with non-evident priors, potentially allowing them to improve performance through prior-based shortcuts rather than image-grounded reasoning.

To probe this possibility, we conduct full-set ablations using Qwen3.5-9B model. We selectively remove the image and/or the scenario name while keeping the question format unchanged. This design separates the contributions of scenario information and visual evidence.
\begin{table}[t]
\centering
\small
\caption{Ablation study of scenario priors on the full 193K MCVQA set, where $^{*}$ denotes the original setting used in the main evaluation.}

\label{tab:scenario_prior_modes}
\setlength{\tabcolsep}{4.5pt}
\renewcommand{\arraystretch}{1.12}
\begin{tabular}{>{\raggedright\arraybackslash}p{0.68\linewidth} c}
\toprule
\textbf{Input mode} & \textbf{Acc.} \\
\midrule
Question only & 29.26\% \\
Question + Scenario, without image & 39.17\% \\
\textbf{Question + Image, without Scenario} & \textbf{64.13\%} \\
\textbf{Question + Image + Scenario$^{*}$} & \textbf{64.89\%} \\
\bottomrule
\end{tabular}
\end{table}
As shown in Table~\ref{tab:scenario_prior_modes}, explicitly providing the scenario name yields a measurable prior when images are absent: the accuracy increases from 29.26\% for question-only input to 39.17\% for question-plus-scenario input, a gain of 9.91 percentage points. However, visual evidence is substantially more informative: adding the image to the question-only setting improves accuracy by 34.87 percentage points, from 29.26\% to 64.13\%. Moreover, removing scenario information from the original evaluation setting results in only a 0.76-percentage-point drop (64.89\% to 64.13\%). Therefore, although scenario information provides a measurable prior, the performance in our main evaluation protocol is primarily driven by image-grounded reasoning rather than by exploiting scenario information as a shortcut.


\section{Evaluation Pipeline Algorithms} \label{app:eval_pipeline}

The full procedural steps for the evaluation of vision-language models on the MCVQA Fire dataset are detailed in Algorithm~\ref{alg:vqa-eval}, which outlines how each image is loaded from the provided root directory, pairing with respective scenario labels and question-specific prompts. Besides, it also illustrates how model predictions are generated under a maximum token limit $T_{max}$. The pipeline then extracts answer choices from model outputs to calculate corresponding scores.


\begin{algorithm}[!t]
\caption{Evaluation on MCVQA}
\label{alg:vqa-eval}

\KwIn{
    Vision-language model $\mathcal{M}$, dataset $\mathcal{D}$, image root $\mathcal{I}$, max tokens $T_{\max}$
}
\KwOut{
    Accuracy $\mathrm{Acc}$, per-dimension $\mathrm{Acc}_{\mathrm{dim}}$, per-scenario $\mathrm{Acc}_{\mathrm{scen}}$
}

Initialize accuracy counters $C, N \leftarrow 0$\;

\ForEach{$(s, x) \in \mathcal{D}$}{
    Load image $I \leftarrow \mathcal{I} + s$\;
    
    \If{$I$ unreadable}{
        \textbf{continue}\;
        \textbf{end if}\;
    }

      $sc \leftarrow x[\texttt{scenario}]$ 

    \ForEach{$qa \in x[\texttt{QA\_pairs}]$}{
        Prompt $p \leftarrow \texttt{build\_prompt}(qa, sc)$\;
        Input $z \leftarrow \texttt{processor}(I, p)$\;

        \If{image tokens missing}{
            \textbf{continue}\;
            \textbf{end if}\;
        }

        Output $\hat{y} \leftarrow \mathcal{M}(z)$ with $T_{\max}$\;
        Prediction $\hat{a} \leftarrow$ extract letter from $\hat{y}$\;
        a $\leftarrow$ first letter of $qa[\texttt{answer}]$\;

        $N \leftarrow N + 1$\;

        \If{$\hat{a} = a$}{
            $C \leftarrow C + 1$\;
            increment correct in $\mathrm{Acc}_{\mathrm{dim}},\ \mathrm{Acc}_{\mathrm{scen}}$\;
            \textbf{end if}\;
        }

        Increment total in $\mathrm{Acc}_{\mathrm{dim}},\ \mathrm{Acc}_{\mathrm{scen}}$\;
    }
    \textbf{end for}\;
}
\textbf{end for}\;

$\mathrm{Acc} \leftarrow \frac{C}{N}$\;

\Return{$\mathrm{Acc},\ \mathrm{Acc}_{\mathrm{dim}},\ \mathrm{Acc}_{\mathrm{scen}}$}
\end{algorithm}

For MCVQA scoring, answer extraction is performed on the model's final-answer field rather than by requiring the entire response to be a single letter. Specifically, our prompt asks models to provide the final choice in a machine-readable format, such as \texttt{\textless output\textgreater A\textless/output\textgreater}. For thinking models, reasoning may appear before the final answer, but only the final-answer field is parsed. If the exact field is missing, we additionally recover unambiguous final-choice patterns such as `\texttt{Answer: A}', `\texttt{Option A}', or `\texttt{the option is A}'. Ambiguous or missing final answers are treated as invalid. This post-processing design ensures that the reported accuracy is not a simple, strict single-letter matching.

\begin{algorithm}[!t]
\caption{CLIP-Based Fire Scene Classification}
\label{alg:clip-eval}
\KwIn{Pretrained CLIP model $\mathcal{M}$, dataset $\mathcal{D}$, class names $\mathcal{C}$}
\KwOut{Zero-shot accuracy $\mathrm{Acc}_{zs}$, few-shot accuracy $\mathrm{Acc}_{svm}$, evaluation metrics}

Tokenize prompts $\mathcal{T}_i \leftarrow$ “A photo of a $c_i$” for each $c_i \in \mathcal{C}$\;
Compute normalized text features $F_t \leftarrow \mathcal{M}_\text{text}(\mathcal{T})$\;

\ForEach{class $c$ in $\mathcal{D}$}{
    Load images $I_c$\;
    \ForEach{image $i \in I_c$}{
        Encode and normalize $f_i \leftarrow \mathcal{M}_\text{image}(i)$\;
        Store $f_i$ and label for both evaluation modes\;
    }
    \textbf{end for}\;
}
\textbf{end for}\;

\tcc{Train classification head under different supervision ratios}
\ForEach{ratio $r \in \{1\%, 3\%, 7\%, 15\%, 30\%\}$}{
    Split each class’s features: $r$\% for training, remainder for testing\;
    Train SVM $\mathcal{S}_r$ on training set\;
    Predict labels on test set using $\mathcal{S}_r$\;
    Compute $\mathrm{Acc}_{svm}^{(r)}$, precision, recall, F1\;
    Save results for comparison\;
}
\textbf{end for}\;

Predict zero-shot labels via $\arg\max \text{softmax}(f_i \cdot F_t^\top)$\;
Compute $\mathrm{Acc}_{zs}$ and associated metrics\;
Export all results to Excel\;

\Return{$\mathrm{Acc}_{zs}$, $\mathrm{Acc}_{svm}^{(r)}$, full metric logs}
\end{algorithm}

In Algorithm~\ref{alg:clip-eval}, the full procedural steps for CLIP-based fire scene classification under both zero-shot and linear probe settings are detailed. This algorithm outlines how class-specific textual prompts are tokenized and encoded into normalized text features using the pretrained CLIP text encoder, and indicates the way that image features are extracted and normalized for all samples in the dataset using the CLIP image encoder. Moreover, the algorithm also illustrates how the evaluation across multiple data regimes. Specifically, our algorithm computes zero-shot predictions via cosine similarity between image and text features, and additionally trains a linear SVM classifier on varying fractions (from 1\% to 30\%) of the labeled image features to test linear probe capabilities.

\section{Model Configuration}\label{app:config}
\noindent\textbf{MLLM Configuration.}\quad
We evaluated the following models: \textit{Qwen3.5 (9B, with and without reasoning)}~\cite{qwen3.5}, \textit{GLM-4.1V (9B)}~\cite{hong2025glm}, \textit{Gemma-3 (12B, 27B)}~\cite{gemma_2025}, \textit{Qwen3-VL (32B, with and without reasoning)}~\cite{bai2025qwen3}, \textit{Qwen3.6 (35B-A3B)}~\cite{qwen36_35b_a3b}, and \textit{InternVL3.5 (8B, 38B)}~\cite{wang2025internvl3_5}. Besides, we used \textit{ Mistral-Small-3.1-Instruct (24B) }\cite{mistral_small_3_1_24b_instruct}, \textit{Gemma-3-it (12B)}~\cite{gemma_2025}, and \textit{Llama-3.2-Vision-Instruct (11B) }\cite{llama32_11b_vision_instruct} for majority voting. In the supplementary experiments, we further evaluated \textit{Step3-VL (10B) }\cite{huang2026step3vl10btechnicalreport}, \textit{LLaVA-OneVision-2 (8B) }\cite{an2026llavaonevision2nextgenerationperceptualintelligence}, \textit{Gemma-4-E4B-it} \cite{Gemma-4-E4B-it}, \textit{Molmo2 (8B) }\cite{clark2026molmo2openweightsdata}, \textit{Qwen3.6 (27B)} \cite{qwen3.6-27b}.

All models were obtained from their official Hugging Face repositories under their respective licenses, and evaluated with default inference settings to ensure consistency and comparability across architectures.

\noindent\textbf{Encoder Config.}\quad
For Stage~IV, we evaluate vision-language encoders using both pretrained general encoders and supervised CLIP+SVM baselines. Experiments are conducted on a curated 9.7K-image test subset. The representative pretrained baselines include \textit{CLIP}~\cite{CLIP}, \textit{MetaCLIP2}~\cite{chuang2026metaclip}, \textit{EVA-CLIP}~\cite{sun2023eva} and \textit{OpenCLIP}~\cite{openclipcvpr}.

\section{Bias Analysis on Generator-family Model and Human Performance}
\label{app:qwen_family_bias}

Since the SAFIRE MCVQA annotations were initially generated by Qwen2.5-VL-72B-Instruct, Qwen-family models may be favored by synthetic patterns inherited from the generator. To examine this concern, we additionally expanded the evaluation to include both Qwen-family and non-Qwen-family models under the same prompt, decoding, answer-extraction, and scoring protocol.

We also estimate human reference performance. To obtain a manageable reference estimate, we randomly sampled 700 VQA questions from the full 193K MCVQA set. We recruited six undergraduate volunteers who expressed interest in the project after it was introduced by teaching assistants during regular class sessions. Participation was voluntary, and each participant spent approximately 1.5–2 hours completing the evaluation in a provided laboratory room. No monetary payment or other compensation was provided; therefore, an assessment of whether payment was adequate for participants’ demographics (e.g., country of residence) is not applicable. Three participants completed the 700-question evaluation without additional guidance, while the other three were instructed to pay particular attention to common fire-and-smoke ambiguities, such as distinguishing smoke from water vapor, fog, or haze, and distinguishing flames from streetlights or electric arcs.

  \begin{table}[t]
  \centering
  \small
  \caption{\textbf{Generator-family bias analysis and human reference performance.} Models are evaluated on the
  full MCVQA set, human performance$^{*}$ is evaluated on a randomly sampled 700 MCVQA subset.}
  \label{tab:qwen_family_bias}
  \setlength{\tabcolsep}{4pt}
  \renewcommand{\arraystretch}{1.14}
  \begin{tabular}{>{\raggedright\arraybackslash}p{0.30\linewidth} >{\raggedright\arraybackslash}p{0.48\linewidth} c}
  \toprule
  \textbf{Model group} & \textbf{Model / Mode} & \textbf{Acc.} \\
  \midrule
  \textbf{Non-Qwen family} & Step3-VL-10B & 63.4\% \\
  & InternVL3.5-38B & 62.7\% \\
  & LLaVA-OneVision-2-8B & 61.6\% \\
  & GLM-4.1V-9B-Base & 59.7\% \\
  & Gemma-4-E4B-it & 59.3\% \\
  & Molmo2-8B & 56.9\% \\
  & \textbf{Average} & \textbf{60.6\%} \\
  \midrule
  \textbf{Qwen family} & Qwen3.6-27B non-thinking & 65.8\% \\
  & Qwen3.6-35B-A3B non-thinking & 65.5\% \\
  & Qwen3.5-9B non-thinking & 64.9\% \\
  & Qwen3.5-9B thinking & 62.2\% \\
  & \textbf{Average} & \textbf{64.6\%} \\
  \midrule
  \textbf{Human performance$^{*}$} & Unguided & 81.3\% \\
  & Guided & 83.6\% \\
  & \textbf{Average} & \textbf{82.5\%} \\
  \bottomrule
  \end{tabular}
  \end{table}

  As shown in Table~\ref{tab:qwen_family_bias}, the non-Qwen-family models achieve an average accuracy of 60.6\% on the full SAFIRE MCVQA benchmark, while the Qwen family achieves an average accuracy of 64.6\%, with the best Qwen-family model reaching 65.8\%. This suggests that a generator-family effect may exist, but its magnitude is limited compared with the gap between current MLLMs and human performance. The human study is conducted on a randomly sampled 700-question subset as a reference estimate rather than a strict paired full-set comparison. Nevertheless, the average human accuracy of 82.5\% is substantially higher than both the non-Qwen average and the best Qwen result. This indicates that the human--model gap is much larger than the gap among model families, suggesting that SAFIRE reflects fire-smoke reasoning difficulty for current MLLMs rather than merely generator-family bias.

\section{MLLM Supervised Fine-Tuning Study}
\label{app:mllm_sft}


 The main focus of SAFIRE is benchmark construction and evaluation. Nevertheless, we conduct an additional supervised fine-tuning experiment to examine whether SAFIRE can also provide useful training supervision.
 
 We fine-tune Qwen3-VL-8B-Instruct with LLaMA-Factory using LoRA rather than full fine-tuning. The vision encoder is
  frozen, and LoRA adapters are applied to the language-model projection and feed-forward layers (\texttt{q/k/v/o\_proj} and
  \texttt{gate/up/down\_proj}; rank 16, alpha 32, dropout 0.05). Training uses the \texttt{qwen3\_vl\_nothink} template with
  maximum sequence length 1,024 and maximum image pixels 262,144. We train for one epoch using an effective batch size of 32, learning rate 1e-4 with cosine scheduling and 126 warmup steps, weight decay 0.01, random seed 42, and evaluate the final checkpoint.

\begin{table}[t]
\centering
\small
\caption{\textbf{Supervised fine-tuning result on SAFIRE.} Qwen3-SAFIRE-SFT is initialized from Qwen3-VL-8B and fine-tuned on the image-level training split.}
\label{tab:mllm_sft}
\setlength{\tabcolsep}{5pt}
\renewcommand{\arraystretch}{1.12}
\begin{tabular}{l c c}
\toprule
\textbf{Model} & \textbf{Acc.} & \textbf{Correct / Total} \\
\midrule
Qwen3-VL-8B Baseline & 63.02\% & 12{,}189 / 19{,}340 \\
Qwen3-SAFIRE-SFT & 70.83\% & 13{,}699 / 19{,}340 \\
\midrule
Improvement & +7.81 pp & +1{,}510 correct \\
\bottomrule
\end{tabular}
\end{table}

To avoid information leakage, we split the data at the image level rather than at the question level, ensuring that MCVQAs generated from the same image are not assigned to different splits. The full 193K MCVQA set is split into 70\% training data, 20\% reserved data, and 10\% held-out test data. This gives approximately 135.1K MCVQA samples for supervised fine-tuning and 19,340 MCVQA samples for testing.

As shown in Table~\ref{tab:mllm_sft}, supervised fine-tuning on SAFIRE substantially improves Qwen3-VL-8B, increasing held-out accuracy from 63.02\% to 70.83\% (+7.81 percentage points). In terms of errors, the baseline makes 7,151 mistakes on the held-out set, while Qwen3-SAFIRE-SFT makes 5,641 mistakes, corresponding to a 21.12\% relative reduction in error rate. The improvement is also broad rather than concentrated in a small subset: all 20 scenarios improve, with gains ranging from +5.07 to +10.26 percentage points, and all 10 reasoning dimensions improve, with gains ranging from +1.91 to +16.29 percentage points. These results indicate that SAFIRE is not only useful for evaluating existing MLLMs, but also provides effective supervision for improving fire-smoke reasoning through task-specific fine-tuning.

\section{Visual Diversity Across Fire-Smoke Scenarios}
\label{sec:appendixA}

\begin{figure*}[t!]
    \centering
    \includegraphics[width=\linewidth]{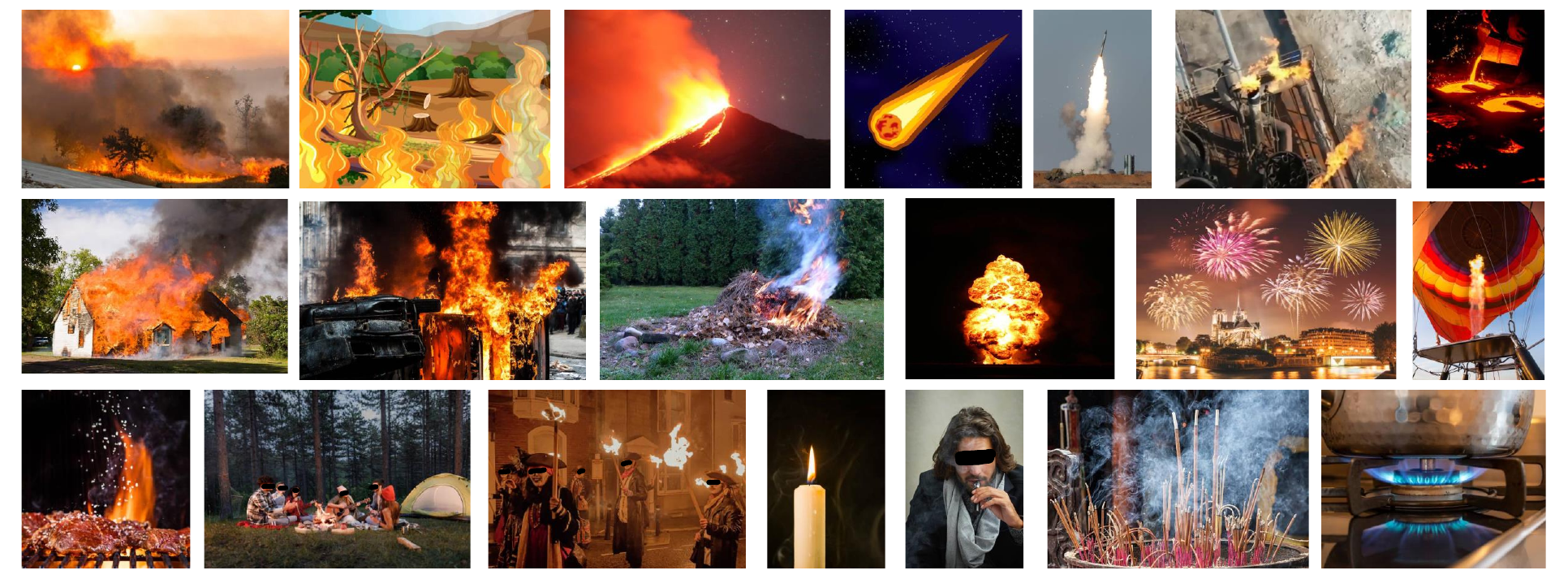}
    \caption{\textbf{Illustration of SAFIRE’s 20 diverse fire and smoke scenarios}, designed to capture both common and uncommon events across natural, civil, industrial, and symbolic contexts. From top-left to bottom-right: (1) Open Grassland Fire, (2) Forest Fire, (3) Volcano, (4) Meteor, (5) Aerospace Operations, (6) Flare Stack, (7) Metal Forging, (8) Residential Scene, (9) Vehicle Fire, (10) Waste Disposal Scene, (11) Explosion, (12) Firework, (13) Hot Air Balloon, (14) Barbecue, (15) Campfire-Bonfire, (16) Torch Parade, (17) Candlelight, (18) Smoking, (19) Incense Burning, (20) Gas Stove.}
    \label{fig:safire_scenarios}
\end{figure*}

\begin{figure*}[!t]
    \begin{subfigure}{\textwidth}
        \centering
        \includegraphics[width=0.95\textwidth]{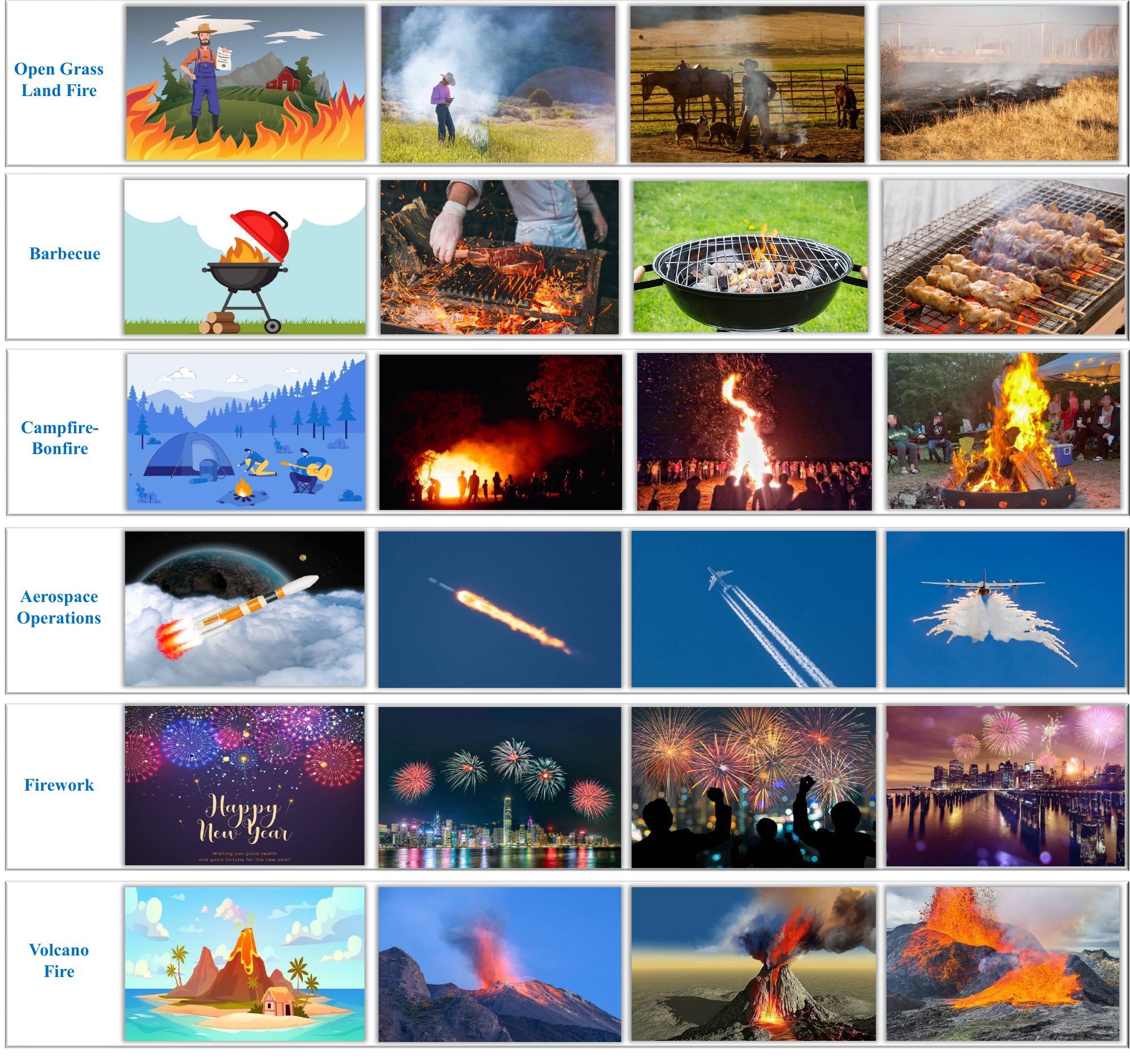}
    \end{subfigure}
    \caption{Visual examples of fire-smoke scenarios used in the benchmark.}
    \label{fig:scenario-1-6}
\end{figure*}

\begin{figure*}[!t]
    \begin{subfigure}{\textwidth}
        \centering
        \includegraphics[width=0.95\textwidth]{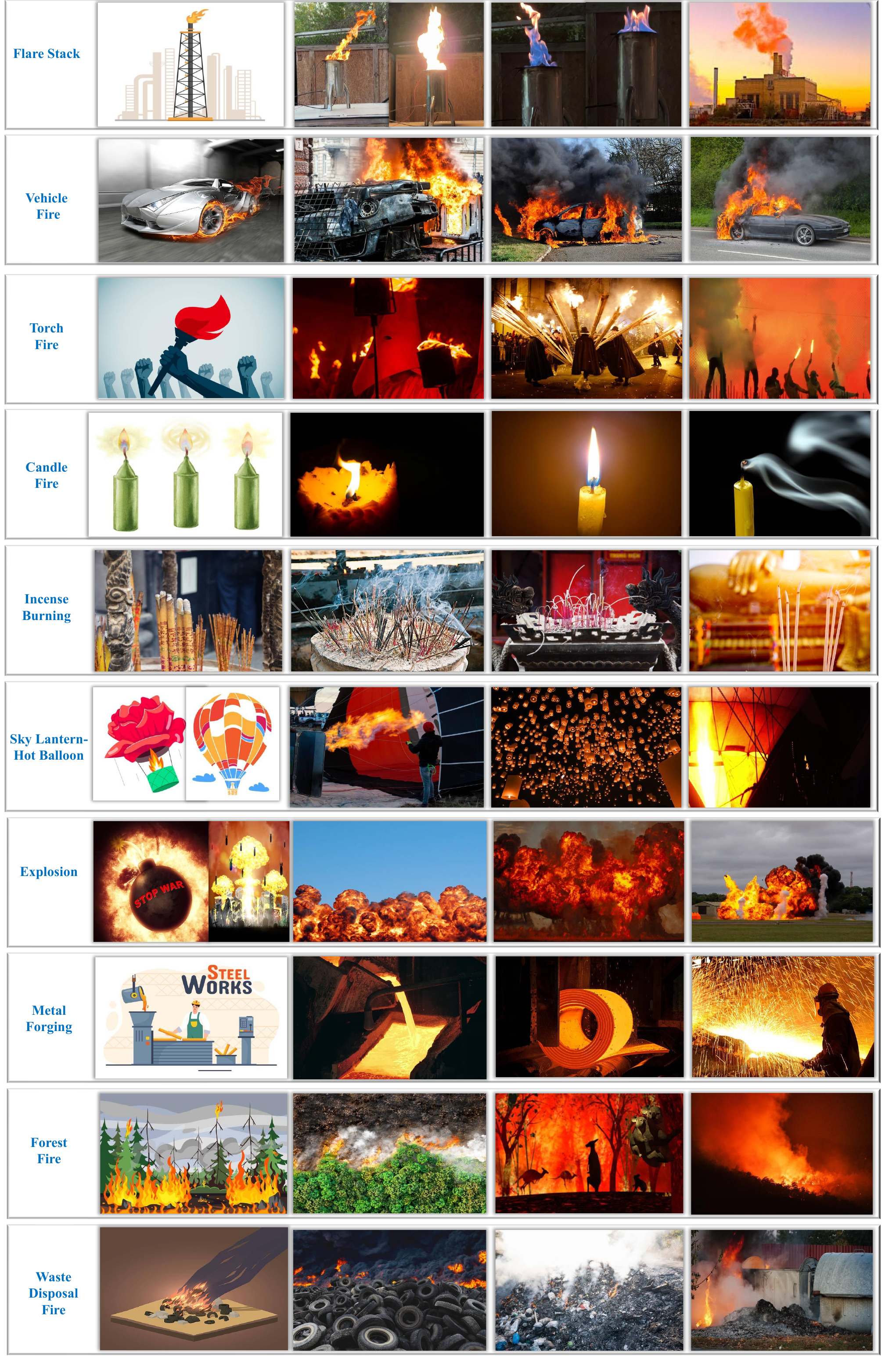}
    \end{subfigure}
 \caption{Visual examples of fire-smoke scenarios used in the benchmark.}
    \label{fig:scenario-9-12}
\end{figure*}

\begin{figure*}[!t]
    \begin{subfigure}{\textwidth}
        \centering
        \includegraphics[width=0.95\textwidth]{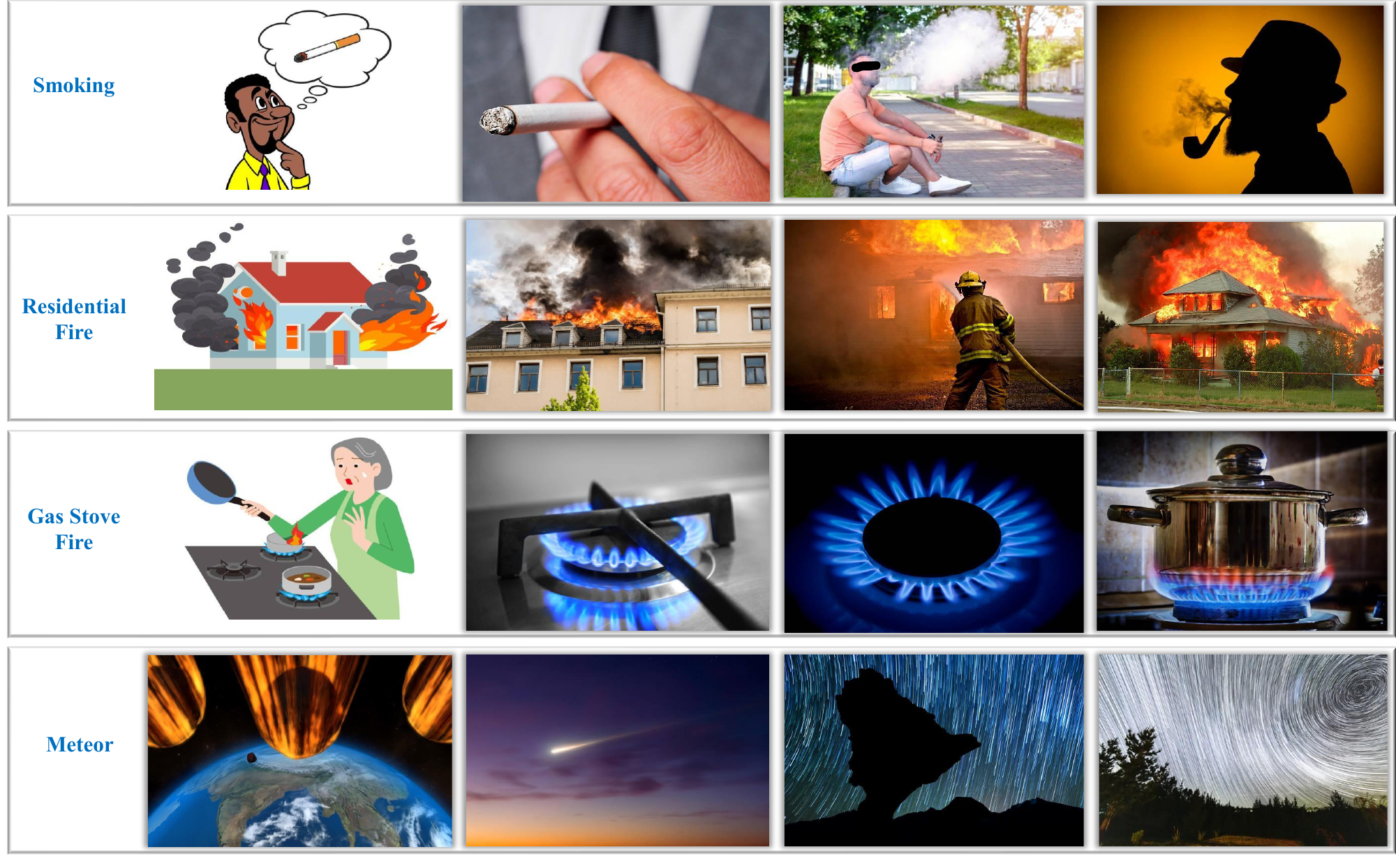}
    \end{subfigure}
    \caption{Visual examples of fire-smoke scenarios used in the benchmark.}
    \label{fig:scenario-17-20}
\end{figure*}

Fig.~\ref{fig:safire_scenarios} illustrates ensemble examples from SAFIRE’s 20 distinct fire and smoke scenarios. While Figures~\ref{fig:scenario-1-6},~\ref{fig:scenario-9-12}, and~\ref{fig:scenario-17-20} showcase 80 additional representative images from the SAFIRE benchmark, covering all 20 fire-smoke scenarios. These visualizations further highlight the dataset’s diversity in fire behavior, smoke characteristics, and contextual settings.

Each scenario includes 4-5 illustrative samples to capture intra-class variation (e.g., perspective, environment, object presence) and inter-class distinctiveness. For example, \textbf{Fig.~\ref{fig:scenario-1-6}} presents large-scale natural fires such as \textit{Volcano} and dynamic man-made events like \textit{Aerospace Operations} and \textit{Burning Vehicles}. \textbf{Fig.~\ref{fig:scenario-9-12}} features culturally embedded and industrial scenarios including \textit{Torch Fire}, \textit{Metal Forging Fire}, and \textit{Explosion}, each with unique flame structures and textures. \textbf{Fig.~\ref{fig:scenario-17-20}} covers smaller or ambiguous sources such as \textit{Smoking}, \textit{Gas Stove Fire}, and \textit{Meteorite Fire}, which are essential for broadening the benchmark beyond conventional scene types.

The scenarios align with the dataset folder structure on our Hugging Face repository, supporting reproducibility and traceability. These extended visual samples emphasize SAFIRE’s focus on realistic, fine-grained fire-smoke phenomena across natural, industrial, accidental, civil controlled, and recreational contexts, and help illustrate the generalization challenges faced by both vision-only and multimodal models.

\section{A Full Generated MCVQA Example}
\label{sec:appendixB}
\begin{figure}[t]
    \centering
    \includegraphics[width=0.85\linewidth]{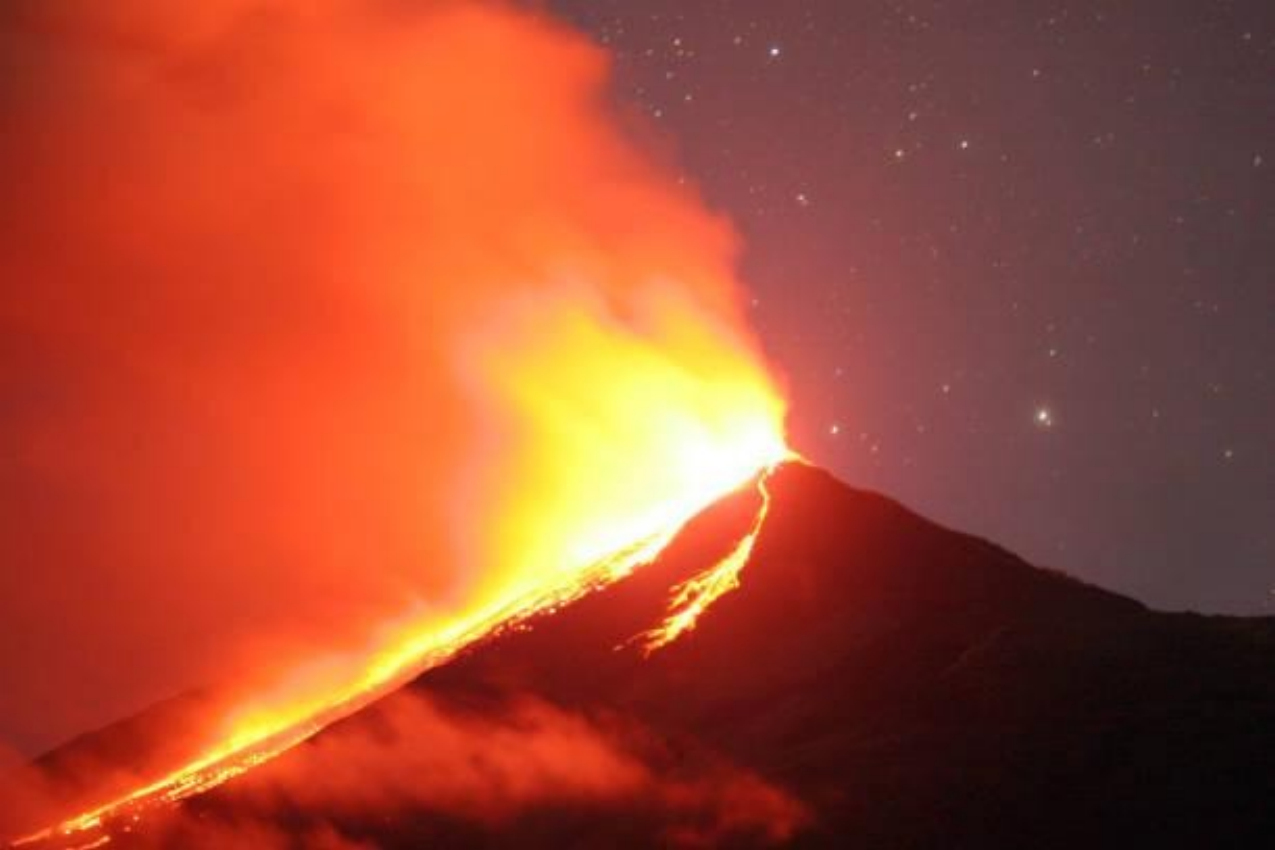}
    \caption{Example image from the Volcano scenario.}
    \label{fig:volcano-img}
\end{figure}

\textbf{Example Annotation Overview.} The image \texttt{volca\_10.jpg} in Fig.\ref{fig:volcano-img} depicts a dramatic volcanic eruption occurring at night.The comprehensive metadata includes a set of 20 multiple-choice visual questions that test various levels of multimodal reasoning. These questions assess quantitative perception (one distinct fire/smoke source), and physical properties (flame color, smoke density and color). Higher-order reasoning questions probe the likely cause of the fire (natural eruption), the absence of human or animal life, and the expected public reaction (evacuation). Moreover, abstract linguistic reasoning is also evaluated, such as interpreting idiomatic expressions like “on fire” and predicting plausible future developments (e.g., lava activity will continue). Details can be found in Figures \ref{fig:mcvqa-json-part1}, and \ref{fig:mcvqa-json-part2}. Overall, we pick out this example here to illustrate how SAFIRE’s MCVQA framework captures both factual visual grounding and inferential logic, providing a rigorous benchmark for assessing the safety-relevant capabilities of modern multimodal models.\\

\clearpage



\begin{figure*}[t]
\begin{tcolorbox}[title=MCVQA JSON Annotation (Part 1),
    colback=gray!5,
    colframe=black,
    boxrule=0.3pt,
    enhanced,
    sharp corners,
    width=\textwidth   
]
\begin{lstlisting}[language=json,
    breaklines=true,           % 自动/手动换行
    breakatwhitespace=false,   % 不仅限空格处
    basicstyle=\small\ttfamily,
    basewidth=0.5em
]
{
  "Image_name": "volca_10.jpg",
  "QA_pairs": [
    {
      "question": "What kind of smoke/flame is this?",
      "options": ["A. Bright lava eruption at night", "B. Geothermal steam vent", "C. Building fire", "D. Fire icon"],
      "answer": "A. Bright lava eruption at night"
    },
    {
      "question": "Where is the fire located in the scene?",
      "options": ["A. Left side", "B. Right side", "C. Center", "D. Not visible"],
      "answer": "C. Center"
    },
    {
      "question": "How many distinct fire or smoke sources are visible?",
      "options": ["A. None", "B. One", "C. Two", "D. More than two"],
      "answer": "B. One"
    },
    {
      "question": "What is the apparent danger level of this fire?",
      "options": ["A. Low", "B. Moderate", "C. High", "D. Extreme"],
      "answer": "D. Extreme"
    },
    {
      "question": "How much smoke is visible in the image?",
      "options": ["A. No smoke", "B. Light", "C. Moderate", "D. Thick smoke"],
      "answer": "D. Thick smoke"
    },
    {
      "question": "What is the color of the smoke?",
      "options": ["A. White", "B. Black", "C. Gray", "D. No smoke"],
      "answer": "C. Gray"
    },
    {
      "question": "Are flames visible?",
      "options": ["A. Yes", "B. No", "C. Not sure", "D. Not applicable"],
      "answer": "A. Yes"
    },
    {
      "question": "What is the color of the flames?",
      "options": ["A. Orange/Yellow", "B. Red", "C. Blue/White", "D. No flames"],
      "answer": "A. Orange/Yellow"
    },
     {
      "question": "What object or material is burning?",
      "options": ["A. Lava or volcanic material", "B. Vegetation", "C. Nothing is visibly burning", "D. Not clear"],
      "answer": "A. Lava or volcanic material"
    },
     {
      "question": "Is this a real fire scene or just an icon or sketch?",
      "options": ["A. Real fire", "B. Icon/Symbol/sketch", "C. Simulation", "D. Not sure"],
      "answer": "A. Real fire"
    },
    {
      "question": "What likely caused the situation?",
      "options": ["A. Natural volcanic eruption", "B. Graphic design", "C. Industrial process", "D. Unknown"],
      "answer": "A. Natural volcanic eruption"
    },

\end{lstlisting}
\end{tcolorbox}
 \caption{A generated MCVQA annotation example from the Volcano scenario, Part 1.}
  \label{fig:mcvqa-json-part1}
\end{figure*}

\begin{figure*}[t]
\begin{tcolorbox}[title=MCVQA JSON (Part 2),
    boxrule=0.3pt,
    enhanced,
    sharp corners,
    width=\textwidth   
]
\begin{lstlisting}[language=json,
    breaklines=true,           % 自动/手动换行
    breakatwhitespace=false,   % 不仅限空格处
    basicstyle=\small\ttfamily,
    basewidth=0.5em
]
    {
      "question": "Are there people or animals in the image?",
      "options": ["A. No", "B. People present", "C. Animals present", "D. Not sure"],
      "answer": "A. No"
    },
    {
      "question": "Is the fire being actively controlled?",
      "options": ["A. Yes", "B. No", "C. Not applicable", "D. Cannot tell"],
      "answer": "B. No"
    },
    {
      "question": "What is the size of the fire?",
      "options": ["A. Small", "B. Medium", "C. Large", "D. Massive"],
      "answer": "D. Massive"
    },
    {
      "question": "What time of day does the image suggest?",
      "options": ["A. Daytime", "B. Nighttime", "C. Indoor lighting", "D. Not determinable"],
      "answer": "B. Nighttime"
    },
    {
      "question": "What is the spatial relationship between the smoke and the fire?",
      "options": ["A. Smoke above the fire", "B. Smoke to the left of Fire", "C. Smoke to the right of Fire", "D. Not applicable"],
      "answer": "A. Smoke above the fire"
    },
    {
      "question": "What will happen next to the Fire/Smoke?",
      "options": ["A. Eruption or lava activity continues", "B. Smoke disperses", "C. Fire quenches", "D. Explosion occurs"],
      "answer": "A. Eruption or lava activity continues"
    },
    {
      "question": "What is the purpose of this fire?",
      "options": ["A. Natural volcanic eruption", "B. Industrial operation", "C. Celebration", "D. Ritual use"],
      "answer": "A. Natural volcanic eruption"
    },
    {
      "question": "How would people react to this smoke/fire?",
      "options": ["A. Take photos only", "B. Ignore it", "C. Evacuate the area", "D. Watch from nearby"],
      "answer": "C. Evacuate the area"
    },
    {
      "question": "What does 'on fire' mean here?",
      "options": ["A. Literal fire", "B. Excited situation", "C. Under attack", "D. Figurative speech"],
      "answer": "A. Literal fire"
    }
  ]
}
\end{lstlisting}
\end{tcolorbox}
\caption{A generated MCVQA annotation example from the Volcano scenario, Part 2.}
\label{fig:mcvqa-json-part2}
\end{figure*}

\end{document}